\documentclass[10pt,twocolumn,letterpaper]{article}

\usepackage[pagenumbers]{cvpr} 

\definecolor{cvprblue}{rgb}{0.21,0.49,0.74}
\usepackage[pagebackref,breaklinks,colorlinks,allcolors=cvprblue]{hyperref}

\def\confName{CVPR}
\def\confYear{2027}
\usepackage{amsmath,amssymb,amsfonts}
\usepackage{booktabs}
\usepackage{multirow}
\usepackage{array}
\usepackage{tabularx}
\usepackage{graphicx}
\usepackage{xspace}
\usepackage{microtype}
\usepackage[T1]{fontenc}
\usepackage[utf8]{inputenc}
\usepackage{enumitem}
\usepackage{booktabs}
\usepackage{graphicx}
\usepackage{makecell}
\usepackage[table]{xcolor}

\definecolor{wizardblue}{RGB}{235,242,251}
\usepackage{pifont}

\usepackage{CJKutf8}

\definecolor{myblue}{RGB}{161,194,250}
\definecolor{mygreen}{RGB}{51,168,83}
\definecolor{myyellow}{RGB}{251,188,3}
\definecolor{myred}{RGB}{234,67,53}
\definecolor{mygrey}{RGB}{95,99,104}
\definecolor{mypup}{RGB}{153,0,204}
\definecolor{mypink}{RGB}{255,182,193}

\title{
    \makebox[\textwidth][c]{
        \raisebox{-0.3cm}{
            \includegraphics[height=1.5cm]{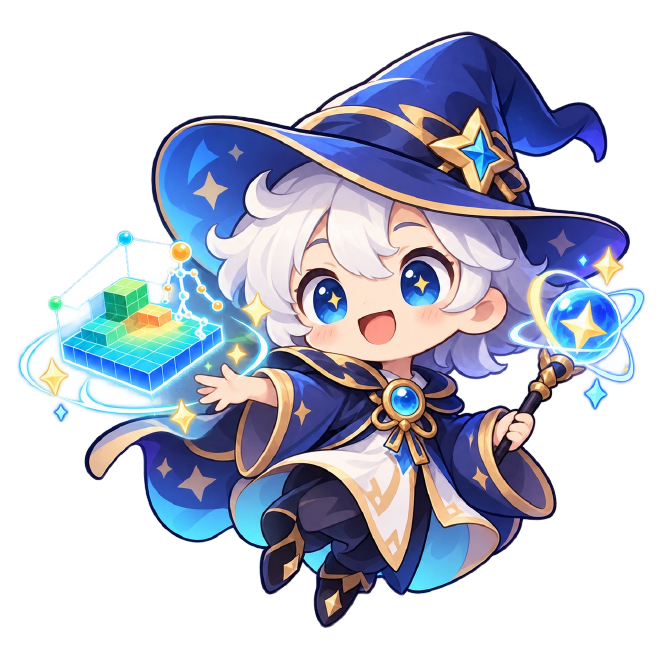}
        }
        \hspace{-0.35cm}
        \mbox{Wizard: Manipulate Any World in Real Time via Agentic Control}
    }
}

\author{%
    Yuzhe Wu$^{1,2,*}$
    \quad Zijun Lin$^{1,3,*}$
    \quad Cheston Tan$^{4}$
    \quad Bihan Wen$^{3}$
    \quad Yeying Jin$^{1,2,\diamondsuit}$\\[0.4em]
    $^{1}$ Tencent
    \quad $^{2}$ NUS
    \quad $^{3}$ NTU
    \quad $^{4}$ A*STAR
}

\begin{document}
\twocolumn[{%
    \renewcommand\twocolumn[1][]{#1}%
    \maketitle

    \begin{center}
        \includegraphics[width=\textwidth]{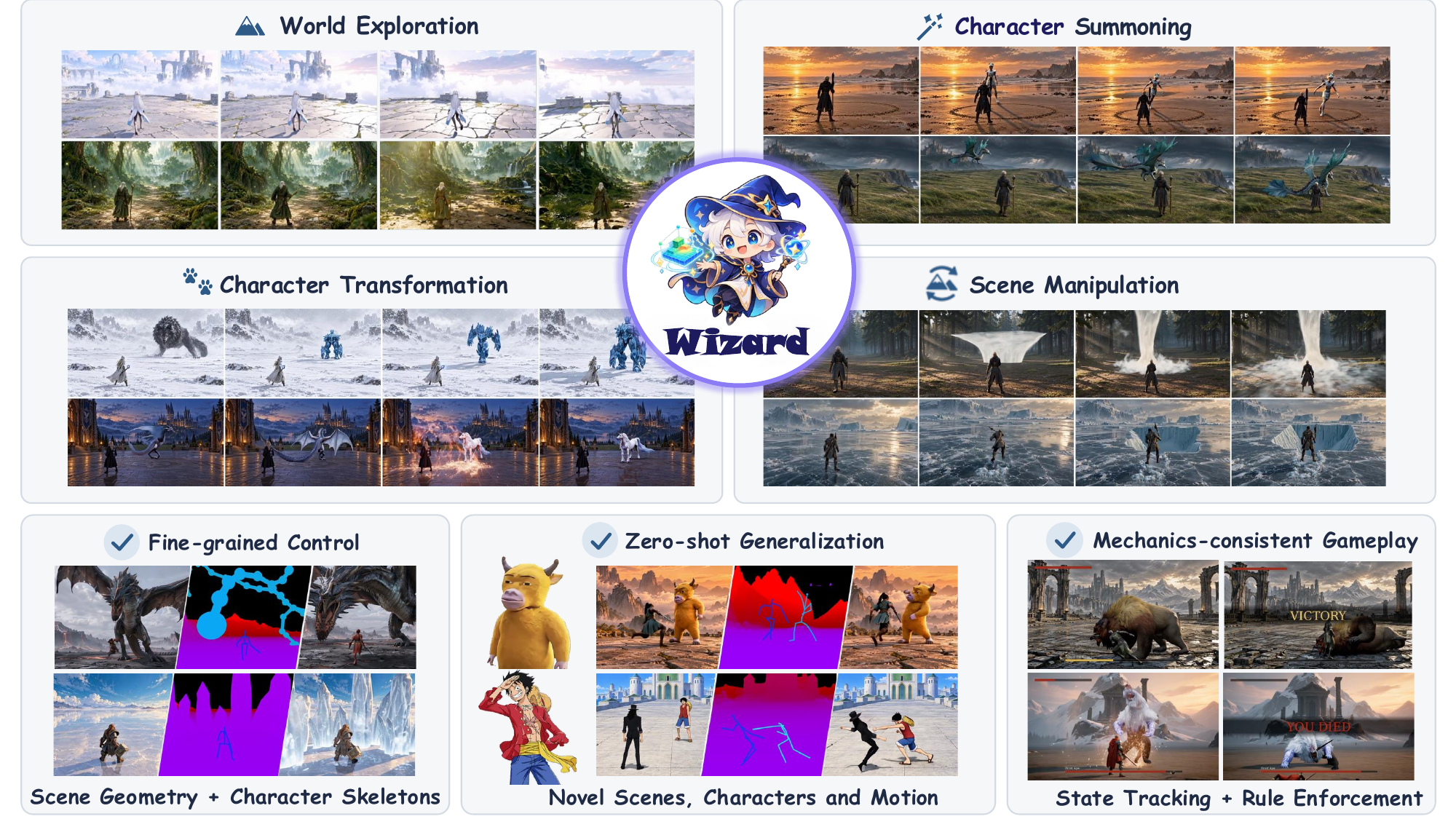}
        \captionof{figure}{\textbf{Overview of Wizard}. Our method supports diverse gameplay experiences, including world exploration, character summoning, character transformation, and scene manipulation. By combining agentic control with geometric scene proxies and articulated character skeletons, Wizard achieves fine-grained control, zero-shot generalization, and mechanics-consistent gameplay with \textbf{real-time interaction}.}
        \label{fig:teaser}
    \end{center}
}]

\begingroup
\renewcommand{\thefootnote}{\ensuremath{*}}
\footnotetext[0]{Equal Contribution.}

\renewcommand{\thefootnote}{\ensuremath{\diamondsuit}}
\footnotetext[0]{Corresponding Author and Project Leader.}
\endgroup

\begin{abstract}
Recent advances in video world models have enabled increasingly realistic and interactive virtual environments, opening new possibilities for player-customizable gameplay. Agentic frameworks further advance this direction by decoupling explicit game logic and controllable world representations from visual generation. However, existing approaches support only simple game settings and predefined motion patterns, generalize poorly to unseen environments and characters, and offer limited real-time interactivity. In this paper, we introduce \textbf{Wizard}, an agentic game world model that enables real-time, precise control over character motions while generalizing zero-shot across novel scenes, characters, and actions. To achieve this, we construct a diverse dataset spanning 805 scenes and 178 characters, incorporating skeleton and geometry information as structured proxies to condition video generation. Our agentic framework explicitly tracks and updates game states to support character-aware adaptive masking and world-anchored spatial conditioning, improving character identity preservation and spatial consistency. Experiments on WBench show that our method achieves higher gameplay fidelity with comparable video quality and spatiotemporal consistency, while using fewer parameters and less training data than state-of-the-art world models. These advantages are also retained in our two-step real-time variant. Overall, Wizard combines agentic control with structured proxy-based generation to enable player-driven game creation with real-time, generalizable, and mechanically consistent world simulation. Project Page: \href{https://wyzardwu.github.io/wizard/}{\textcolor{cvprblue}{https://wyzardwu.github.io/wizard/}}.
\end{abstract}

\section{Introduction}
\label{sec:intro}

World models aim to predict future observations from visual histories and current actions, attracting growing interest in gaming applications \cite{sun2025worldplay, bruce2024genie,he2025matrix,mao2026yume1,gao2026infinite}. By conditioning video generation on players’ mouse and keyboard inputs, video world models offer a natural foundation for interactive and personalized gameplay \cite{huang2026multics,yu2026worldcrafter,zhu2026sana, ye2026worldcast,lin2026stateplay}. 

Despite recent progress, existing game world models face limitations in control flexibility, generalization, and real-time interaction. Action-conditioned models \cite{wang2026reactivegwm, tong2026scope, huang2026solarwm} typically restrict players to predefined action sets, while prompt-conditioned models \cite{zhu2026incantation, deng2026worldmind,chen2026h3} often depend on motion and skill patterns seen during training, limiting their ability to follow unseen instructions. 

To improve state consistency and spatial controllability, recent agentic frameworks maintain explicit world states \cite{lin2026gamedirector, deng2026worldmind} and translate them into structured proxies, such as depth maps \cite{meng2026marionette}, semantic maps \cite{chen2026code}, blockouts \cite{liu2026anygamerenderer}, or oriented 3D bounding boxes \cite{huang2026programmable}, to guide video generation. However, their demonstrations largely focus on basic movement, such as vehicle driving \cite{streamrenderh3_2026, li2026solvideoinferenceengine} and character locomotion \cite{yang2026oneira,zhang2026matrix}, with limited support for articulated character actions, as their proxies do not explicitly encode joint-level motion. Engine-based approaches \cite{huang2026generativeworldrenderer,zhan2026magpie} offer another route to structured control, but their reliance on predefined scenes and additional processing limits scene flexibility and increases latency.

To address this gap, we introduce \textbf{Wizard} (Fig.~\ref{fig:teaser}), an agentic game world model that combines \textbf{fine-grained character control, zero-shot generalization, real-time interaction and mechanically consistent gameplay}, enabling players to customize their game worlds with novel scenes, characters, and motions specified through reference images or free-form language. At its core, Wizard employs an agentic framework to decouple scene construction, character animation, and gameplay logic from visual generation through structured spatial conditioning.

To enable fine-grained camera and character control, we construct a unified visual proxy (Fig.~\ref{fig:teaser}) comprising scene geometry that encodes depth and height, and articulated character skeletons. We reconstruct an approximate traversable 3D environment from a user-provided scene image and convert character references into controllable articulated skeletons, with channel coding distinguishing character instances and their camera-relative orientations. The agent renders view-dependent geometric maps and animates skeletons according to camera and action inputs, providing precise spatial and motion conditioning for the causal video generator. For novel characters, a vision-language model retrieves compatible skeletons and animations based on morphology and intended motion style from a high-quality motion library, then selected motions are retargeted to support diverse, realistic movements. The agent also tracks health, relative distance, and orientation to enforce gameplay rules and adapt boss behavior to evolving combat situations.


In addition, we specifically construct 56.1 hours of aligned proxy–video pairs from diverse sources, including data specifically designed for skeletal control, to train the video model to faithfully follow structured proxies and generalize across scenes, characters, and motions. We further introduce \textbf{character-aware adaptive masking} that restricts reference-image attention to the corresponding character regions, strengthening appearance alignment and preserving character identity during motion. 

To improve spatial consistency, we propose \textbf{world-anchored spatial conditioning} that uses agent-maintained exploration states to distinguish known and unknown regions in the proxies and injects per-pixel world coordinates only into known regions, preserving consistency in previously generated areas while allowing the model to freely synthesize unexplored content. Finally, we apply few-step distillation \cite{zhao2026causal} to enable real-time video generation at 20 FPS, bringing detailed proxy control and consistent visual rendering into an interactive gameplay loop.


\begin{figure*}[t]
\centering
\includegraphics[width=\linewidth]{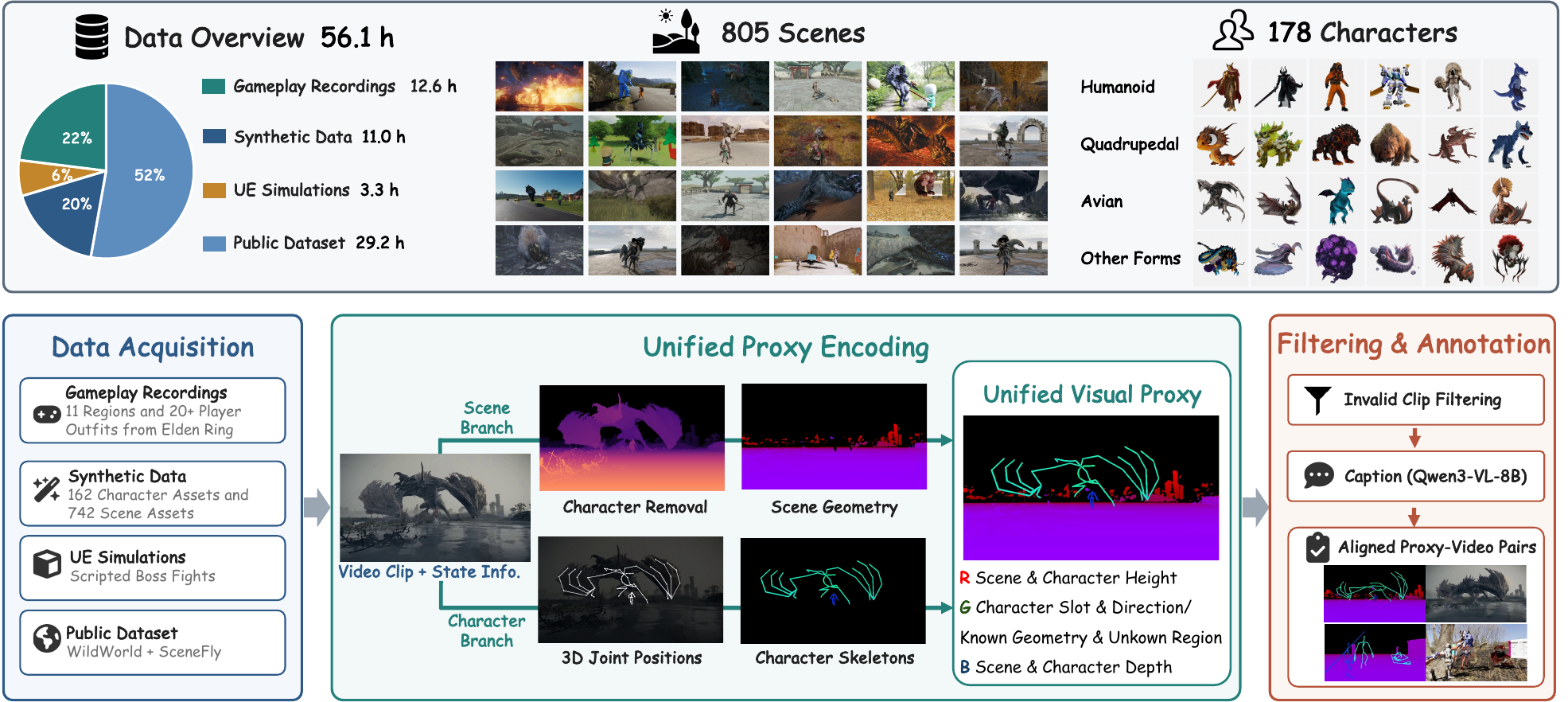}
\caption{\textbf{Dataset Overview and Processing Pipeline.} We construct 56.1 hours of data spanning 805 scenes and 178 characters from four complementary sources. Scene geometry and character skeletons are jointly encoded into unified visual proxies, followed by filtering and captioning to produce aligned proxy--video training pairs.}
\label{fig:data}
\end{figure*}

In summary, our contributions are as follows.
\begin{itemize}[leftmargin=*,topsep=3pt,itemsep=2pt]
\item We propose \textbf{Wizard}, an agentic game world model that enables real-time, zero-shot generalization across unseen scenes, characters, and motions with precise control, substantially expanding the possibilities for player-driven game creation.
\item We build a diverse dataset covering 805 scenes and 178 characters from public videos, game footage, and synthetic sources, with aligned character skeleton and scene geometry to enhance model generalization.
\item We introduce an agentic framework that leverages structured visual proxies to enable accurate character control and mechanics-consistent gameplay.
\item We leverage character positions in visual proxies and agent-maintained exploration states to enable character-aware adaptive masking and world-anchored spatial conditioning, improving reference-to-video identity alignment and spatial consistency in explored regions while allowing flexible generation of unseen content.
\end{itemize}

\section{Related Work}
\label{sec:related}

\subsection{Video World Model}
Recent advances in video generation have enabled action-conditioned world models that allow users to interactively explore generated environments \cite{bruce2024genie,mao2026yume1,he2025matrix,deng2026actionrope}. In games, these models translate keyboard and mouse inputs into dynamic gameplay, with emerging capabilities including NPC control in ReactiveGWM \cite{wang2026reactivegwm}, fine-grained FPS interactions in SCOPE \cite{tong2026scope}, and multiplayer experiences in WorldCast \cite{ye2026worldcast} and Multics \cite{huang2026multics}. However, relying primarily on pixel-level supervision makes it difficult to maintain persistent game states, retain long-term history, and preserve spatial consistency \cite{huang2026programmable}. To address these limitations, agentic frameworks such as WorldMind \cite{deng2026worldmind} and Marionette \cite{meng2026marionette} decouple explicit state management and gameplay logic from visual generation. Despite these advances, existing methods often remain constrained to the game settings seen during training. In contrast, Wizard combines fine-grained skeletal control with a diverse training dataset to enable zero-shot generalization to unseen scenes, characters, and motions.

\subsection{Agentic World Model}
Agentic world models decouple world-state management and action planning from visual generation, enabling explicit control over world dynamics while video models focus on rendering. Related simulation-guided approaches, such as PhysDreamer \cite{zhang2024physdreamer}, RealWonder \cite{liu2026realwonder}, and RoboRender \cite{huang2026roborender}, leverage explicit physical dynamics to guide video generation. Moving beyond simulator-defined dynamics, GameDirector \cite{lin2026gamedirector} maintains combat states and enforces gameplay rules to support player-configurable game settings, while Code World Model \cite{chen2026code} and Programmable World Model \cite{huang2026programmable} employ agent-generated programs to maintain persistent world states and guide visual generation through structured proxies. However, these methods primarily emphasize state consistency and entity-level control \cite{yang2026oneira, zhang2026matrix}, leaving fine-grained character animation underexplored. Concurrent work, SolarRenderer \cite{fang2026solarrenderer}, enables real-time video rendering from diverse proxies, yet seamlessly integrating this capability into a closed-loop gameplay experience remains a challenge. To bridge this gap, Wizard extends agentic control to scene geometry and articulated skeletons, enabling flexible scene manipulation and precise character animation. A diverse, multi-source training dataset enhances generalization, while causal adaptation and few-step distillation enable real-time video generation for responsive gameplay.
\section{Data}
\label{sec:data}

\begin{figure*}[!t]
\centering
\includegraphics[width=\linewidth]{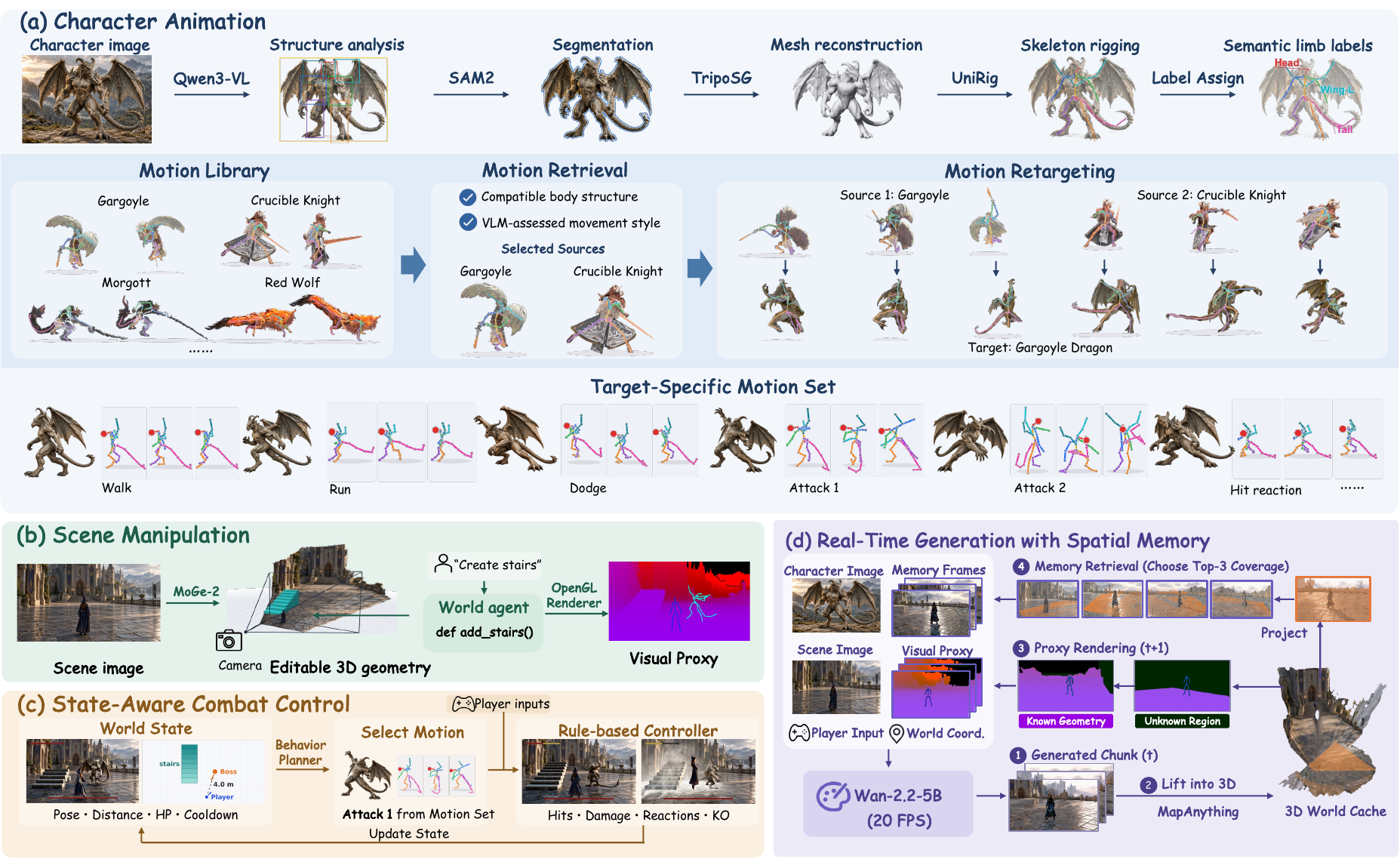}
\caption{\textbf{Inference Pipeline of Wizard.} \textbf{(a) Character animation} is driven by motions retrieved from a motion library and retargeted from compatible source skeletons to the target character’s skeleton. \textbf{(b) Scene manipulation} is performed by a world agent that edits 3D scene geometry and renders visual proxies using OpenGL. \textbf{(c) State-aware combat control} adapts boss behavior to the current game state and enforces explicit mechanics. \textbf{(d) Real-time generation} uses spatial memory with retrieved frames and world coordinates.}
\label{fig:method_overview}
\end{figure*}
Training a video world model with fine-grained control and zero-shot generalization requires diverse visual data paired with complementary proxies tailored to characters and backgrounds, namely articulated skeletons for character motion and geometry maps for scene structure. Existing public datasets for video world models either lack these complementary conditions~\cite{che2025gamegen,yu2025gamefactorly,peng2026compression} or provide them within a single game with limited gameplay diversity~\cite{li2026wildworld}. 

To this end, we construct 56.1 hours of aligned proxy--video pairs spanning 805 scenes and 178 characters. As shown in Fig.~\ref{fig:data}, our dataset combines gameplay recordings, synthetic data, Unreal Engine simulations, and public dataset, covering diverse environments, character morphologies, and motion patterns. Synthetic scenarios further support character entry, multi-character coexistence, disappearance and reappearance, and rapid turns under moving cameras.

For each source, we obtain camera parameters, scene depth, and character skeletons to construct unified visual proxies. Scene geometry and articulated skeletons are jointly rendered under a shared camera, preserving their spatial relationships and mutual occlusion. The RGB channels encode height, character slot and facing direction, and camera-forward depth, respectively. Each character slot is linked to its reference image, decoupling control signals from fixed appearances and supporting generalization to unseen characters. Finally, we filter invalid renders and physically implausible configurations, and use Qwen3-VL-8B \cite{bai2025qwen3} to generate structured captions. Each training sample contains an aligned proxy video, an RGB clip and character reference images. Please refer to Appendix \ref{suppsec:data} for more dataset details.
\section{Method}
\label{sec_method}

As illustrated in Fig.~\ref{fig:method_overview}, Wizard combines an agentic framework for explicit world-state control with a video generator that renders the evolving state through structured visual proxies. Before each gameplay session, an initialization stage taking less than five minutes reconstructs the scene and characters, rigs character skeletons, and builds a target-specific motion set through motion retrieval and retargeting. Once initialized, the framework supports real-time gameplay by updating character motions, scene geometry, and combat states based on player inputs and boss decisions, while the video generator renders the related visual observations. 


\subsection{Agentic Framework}
\label{sec:method_agent}

\paragraph{Character Animation via Skeletal Control.}

Given a user-provided character image, Qwen3-VL~\cite{bai2025qwen3} detects characters and infers their body structures, including posture (upright or horizontal), the number and arrangement of legs, arms, wings, tails, and heads, and their anatomical left--right assignments. We use the predicted bounding boxes to guide SAM2~\cite{ravi2025sam} in segmenting each character, then reconstruct its mesh with TripoSG~\cite{li2025triposg}. UniRig~\cite{zhang2025one} and our medial-axis fitting procedure generate candidate skeletons, from which we select the one that best matches the inferred body structure. We assign semantic labels to its joints and limb chains, identifying body parts such as the left arm or right leg. These labels allow corresponding body parts to be matched across characters with different joint counts and proportions, providing a shared interface for motion retargeting and control.

To enable realistic combat movements for novel characters, we build a reference motion library from CombatMotion~\cite{liao2024animationgpt}, WildWorld~\cite{li2026wildworld}, and animation assets from Elden Ring~\cite{fromsoftware2022eldenring}. For each target character, we retrieve motions from source skeletons with compatible body structures and use a VLM to assess whether the source characters' movement styles, such as aggression, grace, or power, suit the target's appearance. The selected motions can come from multiple sources, allowing each target to inherit a varied moveset. We retarget these motions by matching corresponding limb chains, adapting motion to the target's limb lengths, and applying inverse kinematics to maintain ground contact. We then filter out clips with poor ground contact, body penetration, or unnatural limb crossing. The remaining clips form a motion set tailored to the target character, enabling the controller to retrieve and execute motions in real time without custom animation scripts.

\paragraph{Scene Manipulation via Geometry Control.}
As with character control, users can customize and manipulate scenes according to their preferences. Given a scene image, we use MoGe-2~\cite{wang2026moge} to recover a 3D point map and camera parameters, remove character regions, and construct a background mesh supplemented by a height field. An OpenGL-based offscreen renderer, implemented with ModernGL and EGL, renders this geometry from the current camera viewpoint using the same proxy encoding as in training: the R and B channels encode world height and depth, respectively. As the camera moves, the world agent re-renders the persistent scene geometry to produce a spatially consistent visual proxy. We explain later how the G channel distinguishes visited regions from unknown regions (Fig. \ref{fig:method_overview}(d)).

During gameplay, the world agent, powered by Qwen3-VL-8B, translates users' natural-language instructions into executable programs that compose predefined geometric operators and specify their locations, scales, and deformation parameters. These programs modify the underlying scene geometry, enabling transformations such as raising a platform from the ground or turning the world upside down. The modified geometry is then re-rendered into the proxy video, supporting interactive scene manipulation while maintaining consistent camera projection and spatial structure.

\paragraph{State-Aware Combat Control.}
Our framework maintains a shared world state that tracks both characters’ positions, poses, health, action phases, and skill cooldowns. Based on this state, a behavior planner selects movements and attacks from the previously constructed target-specific motion set. When the player is out of reach, the planner selects a motion that moves the boss toward the player. Within attack range, it evaluates candidate attacks by simulating their swings against the player’s current geometry and prioritizes those likely to connect, subject to action-phase and cooldown constraints. This state-aware planning enables the response to the player’s movements and the evolving combat situation.

The selected motion is then executed in the simulation, where a rule-based controller enforces gameplay constraints and resolves combat outcomes for both characters. It detects hits by checking whether the space swept by an attack intersects the target’s body, excluding periods when the target cannot be hit, such as during a dodge. Valid hits reduce the target’s health, update its health bar, and trigger hit reactions. Skills become available only after their cooldowns expire, and combat ends when either character’s health reaches zero. The updated world state informs the planner’s next decision and is rendered into visual proxies that condition the video generator. Together, state-aware planning and rule-based execution support responsive, mechanics-consistent combat independently of video generation.

\begin{figure*}[!t]
\centering
\includegraphics[width=\linewidth]{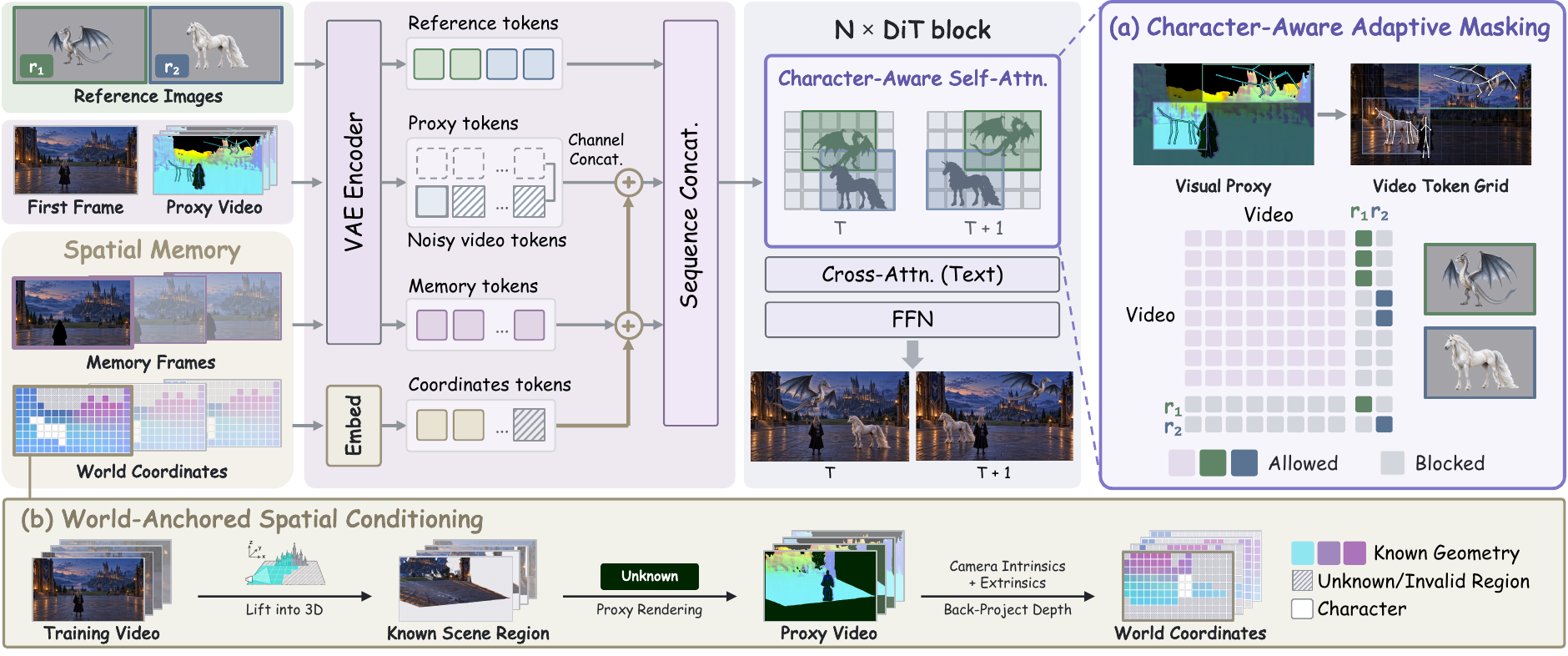}
\caption{\textbf{Model Training.} \textbf{(a) Character-aware adaptive masking} aligns generated characters with their references, while \textbf{(b) world-anchored spatial conditioning} anchors memory and proxy tokens to a shared spatial frame. During inference, the agent maintains the corresponding masks and world coordinates to support consistent generation.}
\label{fig:method_ref}
\end{figure*}

\paragraph{Real-Time Generation with Spatial Memory.}
As illustrated in Fig.~\ref{fig:method_overview}(d), the video generator takes character images, a scene image, and visual proxies to generate video chunks at 20 FPS. Generated frames are lifted into 3D using MapAnything \cite{keetha2025mapanything} to update a persistent world cache, which guides subsequent proxy rendering and memory retrieval.

To enable exploration beyond the input view, the proxy marks unknown regions with a reserved color in the G channel and provides no world coordinates there, allowing the generator to synthesize unseen scenery. Once generated and reconstructed, these regions become known geometry, with height and depth encoded in the R and B channels. This design preserves previously observed geometry while supporting open-ended generation. For memory retrieval, we select up to three historical frames that best cover the region visible from the current viewpoint. These frames, together with the updated proxies, character and scene images, and world coordinates, condition next-chunk generation. World-coordinate injection is detailed in \cref{sec:method_train}.
\subsection{Model Training}
\label{sec:method_train}
As shown in Fig.~\ref{fig:method_ref}, we build our video generator on Wan2.2-TI2V-5B~\cite{wan2025wan}:
\begin{equation}
\label{eq:conditional_generation}
\hat{\mathbf{X}}_{t:t+L-1}
=
G_{\theta}\!\left(
\mathbf{P}_{t:t+L-1}, \mathbf{I}_{0},
\{\mathbf{R}^{k}\}_{k=1}^{K},
\mathbf{C}; \mathbf{H}_{<t}
\right),
\end{equation}
where $\mathbf{P}_{t:t+L-1}$, $\mathbf{I}_{0}$, and
$\mathbf{R}^{k}$ denote the visual proxy sequence, the initial
scene image, and the reference image for character $k$,
respectively. $\mathbf{C}$ denotes the world-coordinate maps associated with
the current proxy sequence and memory frames, while $\mathbf{H}_{<t}$ denotes the context from
previously generated frames.



\paragraph{Character-Aware Adaptive Masking.}
Character skeletons provide explicit spatial information during both training and inference: training data include ground-truth skeletons, while our agentic framework maintains and updates them at runtime. We leverage this shared structure to bind each reference image to its intended character. Each character reference image is VAE-encoded and appended as tokens with a distinct temporal positional offset. For each frame, we derive a bounding region from each character's skeleton and expand it by 25\% to include its surrounding context. In every DiT self-attention layer, we gate attention from video token $i$ to reference token $j$ using an additive mask:
\begin{equation}
\label{eq:character_masking}
\begin{aligned}
A_{ij}
&=
\operatorname{Softmax}_{j}\!\left(
\frac{\mathbf{q}_{i}^{\top}\mathbf{k}_{j}}{\sqrt{d}}
+ M_{ij}
\right),\\
M_{ij}
&=
\begin{cases}
0, & \mathbf{u}_{i}\in\Omega_{k(j),\tau(i)},\\
-\infty, & \text{otherwise},
\end{cases}
\end{aligned}
\end{equation}
where $k(j)$ identifies the character associated with reference token $j$, $\mathbf{u}_{i}$ and $\tau(i)$ denote the spatial position and frame index of video token $i$, and $\Omega_{k,\tau}$ is character $k$'s expanded region on the token grid at frame $\tau$. 


These spatial masks are updated from the explicit character states maintained by our agentic framework, allowing players to control when and where bosses appear or disappear while directing reference guidance to the corresponding regions. As characters move, the masks follow their positions, reducing cross-character appearance leakage. During character transformation, we additionally block attention to cached tokens associated with the previous identity while preserving access to background history and the first-frame sink. By linking appearance guidance to agent-controlled character states, this strategy supports flexible summoning, removal, transformation, and multi-character gameplay while maintaining scene continuity.
\begin{table*}[t]
  \centering
  \caption{{\textbf{Comparison with existing world models} on WBench~\cite{ying2026wbench} (158 navigation cases) and our Gameplay testset (50 clips). We report parameter counts, training data scale, and denoising steps; ``n/r'' means not reported. \textbf{Bold} and \underline{underline} indicate the best and second-best results. Wizard and its real-time variant (20 FPS) achieve strong gameplay fidelity with competitive video and spatiotemporal quality.}}
  \label{tab:main_results}
  \renewcommand{\arraystretch}{1.15}
  \setlength{\tabcolsep}{5pt}
  \resizebox{\textwidth}{!}{%
    \begin{tabular}{@{}lccc|cc|cc|cc@{}}
      \toprule
      & & & & \multicolumn{2}{c|}{\textbf{Video Quality}}
      & \multicolumn{2}{c|}{\textbf{Spatiotemporal Quality}}
      & \multicolumn{2}{c}{\textbf{Gameplay Fidelity}} \\
      \textbf{Method}
      & \textbf{Params}
      & \textbf{Data}
      & \textbf{Steps}
      & \makecell{Imaging\\Quality $\uparrow$}
      & \makecell{Motion\\Smoothness $\uparrow$}
      & \makecell{Background\\Consistency $\uparrow$}
      & \makecell{Spatial\\Consistency $\uparrow$}
      & \makecell{Interaction\\Adherence $\uparrow$}
      & \makecell{Mechanics\\Consistency $\uparrow$} \\
      \midrule
      YUME 1.5~\cite{mao2026yume1}           & 5B  & $>$400\,h & 50 & 63.3 & \underline{97.0} & 90.3 & 71.4 & 45.3 & 88.0 \\
      Matrix-Game 3.0~\cite{wang2026matrix}  & 5B  & 4.8M clips & 50 & \textbf{70.0} & 95.4 & 85.7 & 80.4 & 0.0 & 15.2 \\
      Marionette (Wan2.2-Fun)~\cite{meng2026marionette} & 5B  & 674\,h & 40 & 61.8 & 94.6 & 83.2 & 64.3 & 50.7 & 46.4 \\
      LingBot-World (Base)~\cite{team2026advancing} & 28B & n/r & 70 & \underline{67.9} & 96.9 & \textbf{96.9} & 67.1 & 37.3 & 78.8 \\
      CWM (MiniMax-H3)~\cite{chen2026code}   & 33B & 5.6\,h & 20 & 64.6 & \textbf{98.0} & \underline{92.3} & \textbf{85.0} & \underline{81.3} & \textbf{96.7} \\
      \midrule
      \rowcolor{wizardblue}
      \textbf{Wizard (Ours)}                 & 5B  & 56.1\,h & 30 & 64.7 & \textbf{98.0} & 89.9 & \underline{83.6} & \textbf{86.7} & 96.1 \\
      \rowcolor{wizardblue}
      \textbf{Wizard (Ours, causal)}        & 5B  & 56.1\,h & 2 & 64.4 & 96.9 & 88.2 & 77.4 & \underline{81.3} & \underline{96.3} \\
      \bottomrule
    \end{tabular}%
  }
\end{table*}

\paragraph{World-Anchored Spatial Conditioning.} Beyond memory frames introduced previously, our spatial memory incorporates world coordinates from an explicit scene map maintained and updated by the agent during exploration. We project this map into the viewpoints of both memory frames and the current proxy video to provide spatial conditioning for the video model. For each location $u$, let $K_t(u)$ indicate whether the corresponding scene point has been observed by time $t$. We retain coordinates only for known locations:
\begin{equation}
C_t(u) =
\begin{cases}
\mathbf{X}_t(u), & K_t(u) = 1, \\
\mathbf{0},     & K_t(u) = 0,
\end{cases}
\end{equation}
where $\mathbf{X}_t(u)$ is the world-coordinate position of the scene point projected to $u$. The coordinate maps are embedded into spatially aligned tokens and added element-wise to the corresponding memory tokens and channel-concatenated proxy--video tokens before sequence concatenation. This shared world-coordinate conditioning anchors memory and current generation to the same persistent map, encouraging consistency in previously observed regions while leaving unexplored regions free for synthesis.



\begin{table}[t]
  \centering
  \caption{\textbf{Ablation Studies.} (a) Character-aware adaptive mask and spatial memory. (b) Spatial memory w/ and w/o world coordinates.}
  \label{tab:consistency_ablation}
  \renewcommand{\arraystretch}{1.15}
  \setlength{\tabcolsep}{6pt}
  \resizebox{\columnwidth}{!}{%
    \begin{tabular}{@{}cc|cc@{}}
      \toprule
      \multicolumn{4}{@{}l}{\textit{(a) Proposed modules}} \\
      \midrule
      \makecell{Character-Aware\\Adaptive Mask}
      & \makecell{Spatial Memory}
      & \makecell{Reference\\Binding $\uparrow$}
      & \makecell{Spatial\\Consistency $\uparrow$} \\
      \midrule
      \ding{55} & \ding{55} & 52.8 & 77.9 \\
      \ding{55} & \ding{51} & 52.6 & \underline{81.0} \\
      \ding{51} & \ding{55} & \textbf{76.4} & 77.6 \\
      \rowcolor{wizardblue}
      \ding{51} & \ding{51} & \underline{73.5} & \textbf{82.1} \\
      \bottomrule
      \addlinespace[8pt]
      \toprule
      \multicolumn{4}{@{}l}{\textit{(b) World coordinates in the spatial memory}} \\
      \midrule
      \multicolumn{2}{@{}l|}{Spatial condition}
      & Revisit Error $\downarrow$
      & Copy Bias $\downarrow$ \\
      \midrule
      \multicolumn{2}{@{}l|}{Memory frames only} & 29.43 & 6.61 \\
      \rowcolor{wizardblue}
      \multicolumn{2}{@{}l|}{\cellcolor{wizardblue}+ World coordinates (Ours)} & \textbf{27.44} & \textbf{2.84} \\
      \bottomrule
    \end{tabular}%
  }
\end{table}

\section{Experiments}
\label{sec:experiments}

\subsection{Experimental Setup}

\paragraph{Implementation.}
We first fine-tune Wan2.2-TI2V-5B~\cite{wan2025wan} using 56.1 hours of clips bidirectionally for 35K steps. Following Causal Forcing++~\cite{zhao2026causal}, we adapt it to causal generation and distill it into a real-time 2-step student via consistency distillation and DMD on its own rollouts. As teacher forcing provides a clean history, we corrupt it with the model's one-step reconstruction of a noised copy, forcing the scene to be recovered from the world-anchored memory. Further implementation details are provided in Appendix \ref{sec:supp_impl}.

\paragraph{Benchmarks and Metrics.}
For fair comparison, we evaluate video and spatiotemporal quality using the metrics and 158 navigation cases provided by WBench~\cite{ying2026wbench}. To assess gameplay fidelity, we construct a Gameplay test set of 50 clips featuring characters generated by Qwen-Image \cite{wu2025qwenimage}. Interaction Adherence measures how accurately generated actions follow control inputs, while Mechanics Consistency evaluates adherence to specified game rules. Please refer to Appendix \ref{sec:supp_bench} and \ref{sec:supp_metrics} for more details on evaluation.

\paragraph{Baselines.}
We compare with end-to-end world models (YUME 1.5~\cite{mao2026yume1}, Matrix-Game 3.0~\cite{wang2026matrix}, LingBot-World~\cite{team2026advancing}) and agentic approaches (CWM~\cite{chen2026code}, Marionette~\cite{meng2026marionette}), each driven through its own control inputs (Appendix \ref{sec:supp_baselines}).
The agentic baselines release only bidirectional models, so our main comparison uses our bidirectional model (30 sampling steps).
We also report our distilled causal model, which runs in real time with 2 steps.

\begin{figure*}[!t]
\centering
\includegraphics[width=\linewidth]{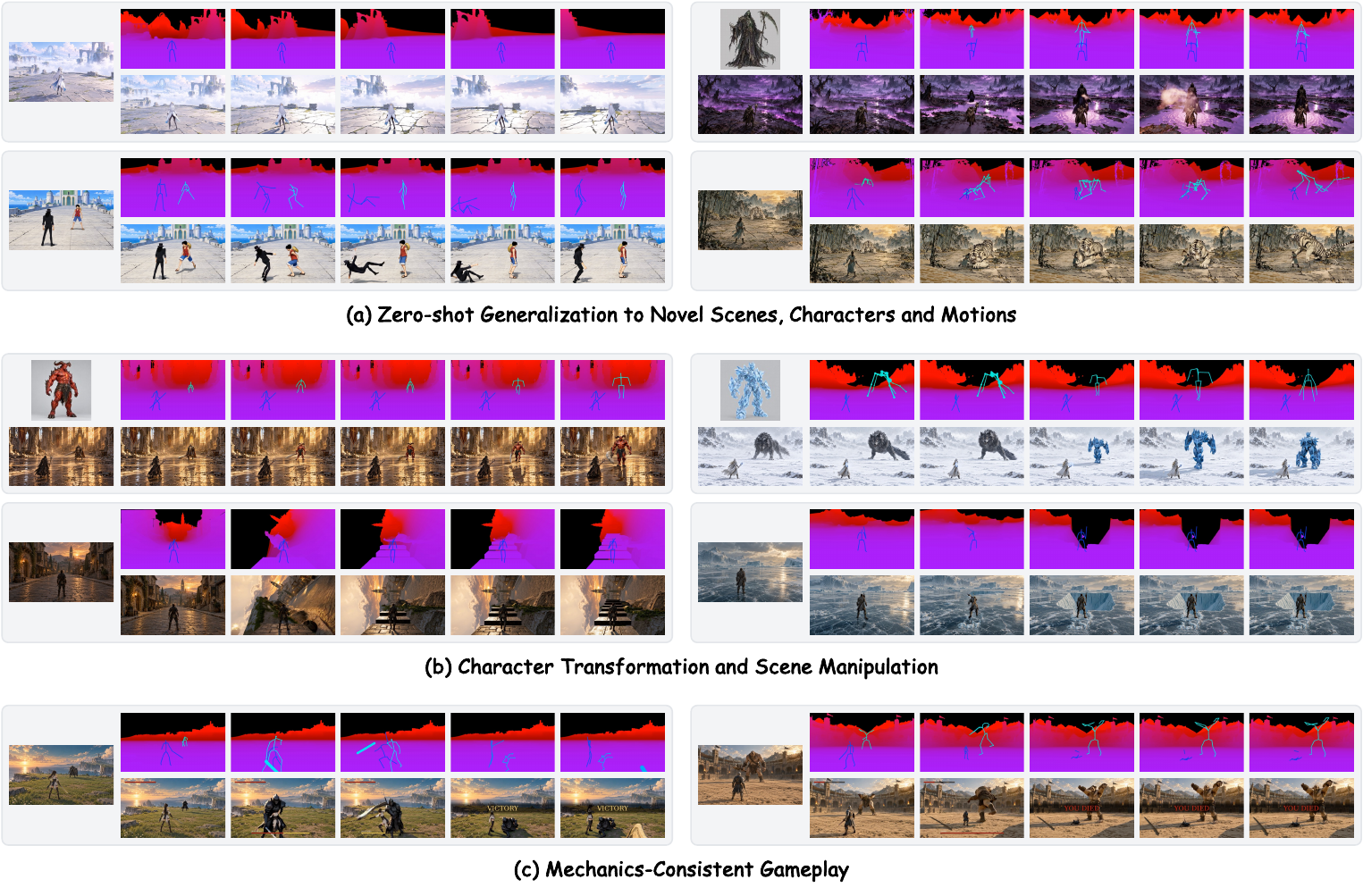}
\caption{\textbf{Qualitative Visualization of Wizard.} For each example, reference images are shown on the left, with visual proxies in the top row and the corresponding video rollouts in the bottom row. Please zoom in for better visibility.}
\label{fig:vis}
\end{figure*}

\subsection{Main Results}
Tab.~\ref{tab:main_results} evaluates general world modeling and gameplay capabilities. On WBench, Wizard achieves competitive video quality with a 5B backbone and 56.1 hours of diverse training data. In particular, the results of motion smoothness highlights the value of articulated skeleton proxies, while spatial memory supports competitive spatiotemporal quality. 

Beyond these general capabilities, agentic control provides strong gameplay fidelity: both Wizard and CWM exceed 95\% mechanics consistency, but our articulated pose control achieves higher interaction adherence than CWM's fixed skeletal pose across actions. These advantages largely persist after distillation, enabling two-step generation at 20 FPS with competitive visual quality and gameplay fidelity. We further demonstrate scene manipulation and character transformation in the qualitative results, as few baselines support the precise zero-shot control required for a meaningful quantitative comparison.

\subsection{Ablation Studies}

\paragraph{Proposed Modules.}
We use 69 \textit{Entry \& Revisit} cases (initial frame from Wbench) to test reference binding between characters with identical skeletons but different references, and spatial consistency after a 150$^\circ$ camera turn and return. In \cref{tab:consistency_ablation}(a), the character-aware adaptive mask raises reference binding from 52.8 to 76.4, while spatial memory improves spatial consistency by up to 4.5 points. Combining both preserves strong character binding and achieves the highest spatial consistency, demonstrating complementary benefits for identity preservation and scene recall.

\paragraph{World Coordinates.}
We evaluate 100 revisit windows from held-out SceneFly scenes~\cite{peng2026compression}, where previously observed regions are revisited from different angles. In \cref{tab:consistency_ablation}(b), adding world coordinates reduces revisit error from 29.43 to 27.44 and copy bias from 6.61 to 2.84. This suggests that world-coordinate conditioning helps re-render visited content from the current viewpoint rather than directly copy the memory frame. Evaluation metric details are provided in Appendix~\ref{sec:supp_metrics}.


\subsection{Qualitative Visualization}
Fig.~\ref{fig:vis} presents unseen user-provided references on the left and paired agent-controlled proxies and generated frames on the right, all using the same model without fine-tuning. Fig.~\ref{fig:vis}(a) shows zero-shot generalization to novel scenes, characters, and motions. Fig.~\ref{fig:vis}(b) shows character transformation and scene manipulation, including environment inversion and new geometry. Fig.~\ref{fig:vis}(c) illustrates mechanics-consistent combat: the agent tracks game states, terminates combat when a character's health reaches zero, and displays the outcome. Together, these examples highlight flexible player control with consistent visual rendering and explicit rule enforcement. Please refer to Appendix~\ref{sec:supp_qual} for qualitative comparisons with baselines and ablated variants.



\section{Conclusions}
\label{sec:conclusion}

In this paper, we introduce \textbf{Wizard}, an agentic game world model that decouples character control, scene manipulation, and gameplay logic from visual generation. Articulated skeletons and geometry maps enable fine-grained character and scene control with mechanics-consistent, real-time interaction. Trained on our diverse dataset, Wizard adopts character-aware adaptive masking and world-anchored spatial conditioning to improve identity preservation and scene consistency. Experiments show strong performance with fewer denoising steps and zero-shot generalization to unseen characters, scenes, and motions. These capabilities enable players to create and manipulate customized game worlds.


{
    \small
    \bibliographystyle{ieeenat_fullname}
    \bibliography{main}

@String(ICCV= {Int. Conf. Comput. Vis.})

@String(TOG= {ACM Trans. Graph.})

@String(ICCV  = {ICCV})

@String(TOG   = {ACM TOG})

@article{wang2026reactivegwm,
  title={ReactiveGWM: Steering NPC in Reactive Game World Models},
  author={Wang, Zeqing and Chen, Danze and Xing, Zhaohu and Tong, Zizhao and Zhang, Yinhan and Yang, Xingyi and Jin, Yeying},
  journal={arXiv preprint arXiv:2605.15256},
  year={2026}
}

@article{tong2026scope,
  title={SCOPE: Simulating Cross-game Operations in Playable Environments for FPS World Models},
  author={Tong, Zizhao and Jin, Yeying and Lai, Hongfeng and Wang, Zeqing and Xing, Zhaohu and Cheng, Kexu and Xu, Haoran and Pu, Zhao and Zhu, Shangwen and Feng, Ruili and others},
  journal={arXiv preprint arXiv:2605.23345},
  year={2026}
}

@article{zhu2026incantation,
  title={Incantation: Natural language as the action interface for multi-entity video world models},
  author={Zhu, Shangwen and Peng, Qianyu and Pu, Zhao and Shu, Zhilei and Ke, Xiangrui and Xing, Zhaohu and Tong, Zizhao and Wang, Zeqing and Cui, Xinyu and Zheng, Zian and others},
  journal={arXiv preprint arXiv:2605.18601},
  year={2026}
}

@article{he2025matrix,
  title={Matrix-game 2.0: An open-source real-time and streaming interactive world model},
  author={He, Xianglong and Peng, Chunli and Liu, Zexiang and Wang, Boyang and Zhang, Yifan and Cui, Qi and Kang, Fei and Jiang, Biao and An, Mengyin and Ren, Yangyang and others},
  journal={arXiv preprint arXiv:2508.13009},
  year={2025}
}

@article{team2026advancing,
  title={Advancing open-source world models},
  author={Team, Robbyant and Gao, Zelin and Wang, Qiuyu and Zeng, Yanhong and Zhu, Jiapeng and Cheng, Ka Leong and Li, Yixuan and Wang, Hanlin and Xu, Yinghao and Ma, Shuailei and others},
  journal={arXiv preprint arXiv:2601.20540},
  year={2026}
}

@article{gao2026infinite,
  title={Infinite Worlds with Versatile Interactions},
  author={Gao, Zelin and Wang, Qiuyu and Zhu, Jiapeng and Chen, Jingye and Liu, Zichen and Bai, Qingyan and Wang, Jiahao and Yuan, Yufeng and Wang, Hanlin and Lu, Yichong and others},
  journal={arXiv preprint arXiv:2607.07534},
  year={2026}
}

@article{li2026wildworld,
  title={Wildworld: A large-scale dataset for dynamic world modeling with actions and explicit state toward generative arpg},
  author={Li, Zhen and Meng, Zian and Shi, Shuwei and Peng, Wenshuo and Wu, Yuwei and Zheng, Bo and Li, Chuanhao and Zhang, Kaipeng},
  journal={arXiv preprint arXiv:2603.23497},
  year={2026}
}

@article{lin2026stateplay,
  title={StatePlay: State-Aware Game World Models for Mechanics-Consistent Generation},
  author={Lin, Zijun and Wang, Zeqing and Tan, Cheston and Wen, Bihan and Jin, Yeying},
  journal={arXiv preprint arXiv:2607.26754},
  year={2026}
}

@article{meng2026marionette,
  title={Marionette: Predicting World States, Rendering Geometry, Painting Appearance},
  author={Meng, Zian and Li, Zhen and Li, Chuanhao and Li, Qiang and Zhang, Kaipeng},
  journal={arXiv preprint arXiv:2608.14530},
  year={2026}
}

@article{deng2026worldmind,
  title={WorldMind: Decoupled Game World Model for State-Aware NPC Behavior},
  author={Deng, Zhiyang and Zhang, Boran and Chen, Danze and Jin, Yeying},
  journal={arXiv preprint arXiv:2608.21439},
  year={2026}
}

@article{chen2026code,
  title={Code World Model: Coding Agent as World Brain},
  author={Chen, Yiwen and Lin, Guosheng and Zhang, Chi},
  journal={arXiv preprint arXiv:2608.25927},
  year={2026}
}

@article{zhan2026magpie,
  title={Magpie: Real-Time World Renderer for Interactive Games},
  author={Zhan, Xiaoyu and Wang, Xinyu and Zhang, Xiaohong and Zhu, Huanjie and Sun, Tengjiao and Fang, Pengcheng and Yu, Jiaxing and Guo, Yanwen and Fu, Dongjie},
  journal={arXiv preprint arXiv:2608.27168},
  year={2026}
}

@article{huang2026programmable,
  title={Programmable World Model},
  author={Huang, Zheng-Hui and Lin, Guixu and Lin, Jiacheng and Huang, Yi-Chuan and Yu, Ruihan and Niu, Muyao and Yang, Siqi and Liu, Yu-Lun and Chuang, Yung-Yu and Zhang, Kaipeng and others},
  journal={arXiv preprint arXiv:2609.10540},
  year={2026}
}

@article{sun2025worldplay,
  title={Worldplay: Towards long-term geometric consistency for real-time interactive world modeling},
  author={Sun, Wenqiang and Zhang, Haiyu and Wang, Haoyuan and Wu, Junta and Wang, Zehan and Wang, Zhenwei and Wang, Yunhong and Zhang, Jun and Wang, Tengfei and Guo, Chunchao},
  journal={arXiv preprint arXiv:2512.14614},
  year={2025}
}

@article{wang2026matrix,
  title={Matrix-game 3.0: Real-time and streaming interactive world model with long-horizon memory},
  author={Wang, Zile and Liu, Zexiang and Li, Jiaxing and Huang, Kaichen and Xu, Baixin and Kang, Fei and An, Mengyin and Wang, Peiyu and Jiang, Biao and Wei, Yichen and others},
  journal={arXiv preprint arXiv:2604.08995},
  year={2026}
}

@inproceedings{bruce2024genie,
  title={Genie: Generative interactive environments},
  author={Bruce, Jake and Dennis, Michael D and Edwards, Ashley and Parker-Holder, Jack and Shi, Yuge and Hughes, Edward and Lai, Matthew and Mavalankar, Aditi and Steigerwald, Richie and Apps, Chris and others},
  booktitle={Forty-first International Conference on Machine Learning},
  year={2024}
}

@article{zhu2026sana,
  title={Sana-wm: Efficient minute-scale world modeling with hybrid linear diffusion transformer},
  author={Zhu, Haoyi and Liu, Haozhe and Zhao, Yuyang and Ye, Tian and Chen, Junsong and Yu, Jincheng and He, Tong and Han, Song and Xie, Enze},
  journal={arXiv preprint arXiv:2605.15178},
  year={2026}
}

@article{wan2025wan,
  title={Wan: Open and advanced large-scale video generative models},
  author={Wan, Team and Wang, Ang and Ai, Baole and Wen, Bin and Mao, Chaojie and Xie, Chen-Wei and Chen, Di and Yu, Feiwu and Zhao, Haiming and Yang, Jianxiao and others},
  journal={arXiv preprint arXiv:2503.20314},
  year={2025}
}

@inproceedings{mao2026yume1,
  title={Yume1.5: A text-controlled interactive world generation model},
  author={Mao, Xiaofeng and Li, Zhen and Li, Chuanhao and Xu, Xiaojie and Ying, Kaining and Zhang, Kaipeng},
  booktitle={Proceedings of the IEEE/CVF Conference on Computer Vision and Pattern Recognition},
  pages={7752--7761},
  year={2026}
}

@misc{openai2026gpt55,
  title        = {{GPT-5.5} System Card},
  author       = {{OpenAI}},
  year         = {2026},
  month        = apr,
  howpublished = {\url{https://openai.com/index/gpt-5-5-system-card/}},
  note         = {Accessed: 2026-07-28}
}

@article{chung2023unimax,
  title={Unimax: Fairer and more effective language sampling for large-scale multilingual pretraining},
  author={Chung, Hyung Won and Constant, Noah and Garcia, Xavier and Roberts, Adam and Tay, Yi and Narang, Sharan and Firat, Orhan},
  journal={arXiv preprint arXiv:2304.09151},
  year={2023}
}

@article{zhao2026causal,
  title={Causal forcing++: Scalable few-step autoregressive diffusion distillation for real-time interactive video generation},
  author={Zhao, Min and Zhu, Hongzhou and Zheng, Kaiwen and Zhou, Zihan and Yan, Bokai and Li, Xinyuan and Yang, Xiao and Li, Chongxuan and Zhu, Jun},
  journal={arXiv preprint arXiv:2605.15141},
  year={2026}
}

@article{chen2026h3,
  title={H3-World: Turning Language Understanding into World Control},
  author={Chen, Danze and Wang, Zeqing and Lin, Ziyue and Yang, Xingyi and Jin, Yeying},
  journal={arXiv preprint arXiv:2609.01560},
  year={2026}
}

@article{lin2026gamedirector,
  title={GameDirector: Decoupling Gameplay Logic from Rendering for Player-Configurable Game World Models},
  author={Lin, Zijun and Deng, Zhiyang and Wu, Yuzhe and Wen, Bihan and Jin, Yeying},
  journal={arXiv preprint arXiv:2609.25652},
  year={2026}
}

@article{ye2026worldcast,
  title={WorldCast: Distributed Multiplayer World Models},
  author={Ye, Ziyang and Huang, Junchao and Zhang, Evelyn and Xie, Zhihao and Zhang, Ruicheng and Han, Boyao and Ban, Litao and Wang, Ziye and Hu, Xinting and Shi, Shaoshuai and others},
  journal={arXiv preprint arXiv:2610.12412},
  year={2026}
}

@misc{streamrenderh3_2026,
  title        = {StreamRender-H3: Interactive Coding Worlds with Real-time Streaming Diffusion Render},
  author       = {Yanzuo Lu and Tian Ye and Yitong Li and Shuchen Xue and Haozhe Liu and Song Han and Enze Xie},
  year         = {2026},
  howpublished = {\url{https://nvlabs.github.io/Sana/Sol-Engine/StreamRender-H3/}},
  note         = {Project page}
}

@misc{li2026solvideoinferenceengine,
  title         = {Sol Video Inference Engine: Agent-Native Full-Stack Acceleration Framework for Efficient Video Generation},
  author        = {Yitong Li and Junsong Chen and Haopeng Li and Haozhe Liu and Jincheng Yu and Ligeng Zhu and Ping Luo and Song Han and Enze Xie},
  year          = {2026},
  eprint        = {2606.23743},
  archivePrefix = {arXiv},
  primaryClass  = {cs.CV},
  doi           = {10.48550/arXiv.2606.23743},
  url           = {https://arxiv.org/abs/2606.23743}
}

@article{huang2026multics,
  title   = {Multics: Generative Multi-Player World Model
             with Hierarchical Communication},
  author  = {Huang, Kaiyi and Yu, Wangbo and Huang, Yukun and
             Yu, Jiwen and Liu, Kunhao and Yang, Donglin and
             Hu, Wenbo and Shan, Ying and Liu, Xihui},
  journal = {arXiv preprint arXiv:2610.17201},
  year    = {2026},
  url     = {https://arxiv.org/abs/2610.17201}
}

@article{yu2026worldcrafter,
  title={WorldCrafter: Consistent Video World Model with Implicit 3D-aware Memory},
  author={Yu, Wangbo and Liu, Kunhao and Hu, Wenbo and Yuan, Shenghai and Feng, Chaoran and Zhou, Haiyang and Huang, Yukun and Wang, Yiran and Zhao, Wang and Luo, Yingmin and others},
  journal={arXiv preprint arXiv:2609.24984},
  year={2026}
}

@article{huang2026solarwm,
  title={SolarWM: Open Data and Scalable Training for Long-Horizon Video World Models},
  author={Huang, Junchao and Fang, Guian and Qian, Shengju and Kong, Xianghao and Zhao, Zhuoran and Huang, Wei and Du, Yihua and Zhang, Zixin and Cui, Justin and Gu, Yuchao and others},
  journal={arXiv preprint arXiv:2609.02886},
  year={2026}
}

@article{huang2026generativeworldrenderer,
    title={Generative World Renderer},
    author={Zheng-Hui Huang and Zhixiang Wang and Jiaming Tan and Ruihan Yu and Yidan Zhang and Bo Zheng and Yu-Lun Liu and Yung-Yu Chuang and Kaipeng Zhang},
    journal={arXiv preprint arXiv:2604.02329},
    year={2026}
}

@article{yang2026oneira,
  title={Oneira: From Open-Ended Generation to Open-World Interaction in Video World Models},
  author={Yang, Xindi and Li, Baolu and Lee, Liam and Yin, Zhenfei and Zhang, Songxin and Song, Zhuoyang and Jia, Xu and Cai, Jianfei and Wong, Tien-Tsin and Jing, Bingyi and Yang, Mengyue},
  journal={arXiv preprint arXiv:2610.01614},
  year={2026}
}

@article{zhang2026matrix,
  title={Matrix-Game 4.0: Harnessing Video Models for Real-Time Playable Game Worlds},
  author={Zhang, Jihai and Qian, Runjia and Kong, Kaiqin and Jiang, Biao and Wang, Puyi and Chen, Siru and Li, Tianle and Chen, Guanjie and Wu, Junyi and Chen, Chubin and others},
  journal={arXiv preprint arXiv:2610.27999},
  year={2026}
}

@inproceedings{zhang2024physdreamer,
  title={Physdreamer: Physics-based interaction with 3d objects via video generation},
  author={Zhang, Tianyuan and Yu, Hong-Xing and Wu, Rundi and Feng, Brandon Y and Zheng, Changxi and Snavely, Noah and Wu, Jiajun and Freeman, William T},
  booktitle={European Conference on Computer Vision},
  pages={388--406},
  year={2024},
  organization={Springer}
}

@article{liu2026realwonder,
  title={Realwonder: Real-time physical action-conditioned video generation},
  author={Liu, Wei and Chen, Ziyu and Li, Zizhang and Wang, Yue and Yu, Hong-Xing and Wu, Jiajun},
  journal={arXiv preprint arXiv:2603.05449},
  year={2026}
}

@article{huang2026roborender,
  title={RoboRender: Robot-Oriented Video Generation for Visual Sim-to-Real Transfer},
  author={Huang, Huang and Ai, Wensi and Chen, Ziyu and Wang, Youhui and Du, Zijian and Liu, Yang and Yang, Jiaolong and Fei-Fei, Li and Wu, Jiajun},
  journal={arXiv preprint arXiv:2610.09254},
  year={2026}
}

@misc{tochilkin2024triposrfast3dobject,
      title={TripoSR: Fast 3D Object Reconstruction from a Single Image}, 
      author={Dmitry Tochilkin and David Pankratz and Zexiang Liu and Zixuan Huang and Adam Letts and Yangguang Li and Ding Liang and Christian Laforte and Varun Jampani and Yan-Pei Cao},
      year={2024},
      eprint={2403.02151},
      archivePrefix={arXiv},
      primaryClass={cs.CV},
      url={https://arxiv.org/abs/2403.02151}, 
}

@inproceedings{che2025gamegen,
  title={Gamegen-x: Interactive open-world game video generation},
  author={Che, Haoxuan and He, Xuanhua and Liu, Quande and Jin, Cheng and Chen, Hao},
  booktitle={International Conference on Learning Representations},
  volume={2025},
  pages={37546--37593},
  year={2025}
}

@inproceedings{yu2025gamefactorly,
  title={GameFactorly: Creating New Games with Generative Interactive Videos},
  author={Yu, Jiwen and Qin, Yiran and Wang, Xintao and Wan, Pengfei and Zhang, Di and Liu, Xihui},
  booktitle={2025 IEEE/CVF International Conference on Computer Vision (ICCV)},
  pages={11590--11599},
  year={2025},
  organization={IEEE}
}

@article{fang2026solarrenderer,
  title={SolarRenderer: Few-Step Generative Reference Rendering},
  author={Fang, Guian and Qian, Shengju and Lu, Dongyue and Huang, Junchao and Jiang, Lutao and Wang, Xin and Song, Yiren and Jiang, Li and Shou, Mike Zheng},
  journal={arXiv preprint arXiv:2610.28630},
  year={2026}
}

@misc{fromsoftware2022eldenring,
  author       = {{FromSoftware}},
  title        = {{Elden Ring}},
  year         = {2022},
  howpublished = {Video game. Bandai Namco Entertainment},
  url          = {https://www.bandainamcoent.com/games/elden-ring}
}

@misc{sketchfab,
  author       = {{Sketchfab}},
  title        = {{Sketchfab}},
  howpublished = {\url{https://sketchfab.com/}},
  note         = {Accessed: 2026-10-16}
}

@misc{quaternius,
  author       = {{Quaternius}},
  title        = {{Quaternius}: Free Game Assets},
  howpublished = {\url{https://quaternius.com/}},
  note         = {Accessed: 2026-10-16}
}

@misc{kaykit,
  author       = {Lousberg, Kay},
  title        = {{KayKit}: Game Assets},
  howpublished = {\url{https://kaylousberg.com/game-assets}},
  note         = {Accessed: 2026-10-16}
}

@misc{polyhaven,
  author       = {{Poly Haven}},
  title        = {{Poly Haven}: The Public 3D Asset Library},
  howpublished = {\url{https://polyhaven.com/}},
  note         = {Accessed: 2026-10-16}
}

@article{bai2025qwen3,
  title={Qwen3-vl technical report},
  author={Bai, Shuai and Cai, Yuxuan and Chen, Ruizhe and Chen, Keqin and Chen, Xionghui and Cheng, Zesen and Deng, Lianghao and Ding, Wei and Gao, Chang and Ge, Chunjiang and others},
  journal={arXiv preprint arXiv:2511.21631},
  year={2025}
}

@inproceedings{ravi2025sam,
  title={Sam 2: Segment anything in images and videos},
  author={Ravi, Nikhila and Gabeur, Valentin and Hu, Yuan-Ting and Hu, Ronghang and Ryali, Chaitanya and Ma, Tengyu and Khedr, Haitham and R{\"a}dle, Roman and Rolland, Chloe and Gustafson, Laura and others},
  booktitle={International Conference on Learning Representations},
  volume={2025},
  pages={28085--28128},
  year={2025}
}

@article{li2025triposg,
  title={Triposg: High-fidelity 3d shape synthesis using large-scale rectified flow models},
  author={Li, Yangguang and Zou, Zi-Xin and Liu, Zexiang and Wang, Dehu and Liang, Yuan and Yu, Zhipeng and Liu, Xingchao and Guo, Yuan-Chen and Liang, Ding and Ouyang, Wanli and others},
  journal={IEEE Transactions on Pattern Analysis and Machine Intelligence},
  year={2025},
  publisher={IEEE}
}

@article{zhang2025one,
  title={One model to rig them all: Diverse skeleton rigging with unirig},
  author={Zhang, Jia-Peng and Pu, Cheng-Feng and Guo, Meng-Hao and Cao, Yan-Pei and Hu, Shi-Min},
  journal={ACM Transactions on Graphics (ToG)},
  volume={44},
  number={4},
  pages={1--18},
  year={2025},
  publisher={ACM, New York, NY, USA}
}

@article{wang2026moge,
  title={Moge-2: Accurate monocular geometry with metric scale and sharp details},
  author={Wang, Ruicheng and Xu, Sicheng and Dong, Yue and Deng, Yu and Xiang, Jianfeng and Lv, Zelong and Sun, Guangzhong and Tong, Xin and Yang, Jiaolong},
  journal={Advances in Neural Information Processing Systems},
  volume={38},
  pages={35928--35959},
  year={2026}
}

@misc{liao2024animationgpt,
  title={AnimationGPT: An AIGC tool for generating game combat motion assets},
  author={Liao, Yihao and Fu, Yiyu and Cheng, Ziming and Wang, Jiangfeiyang},
  year={2024},
  howpublished = {\url{https://github.com/fyyakaxyy/AnimationGPT}}
}

@article{liu2026anygamerenderer,
  title={AnyGameRenderer: Entity-Consistent Generative Rendering Across Games and an Entity-Level Benchmark},
  author={Liu, Ruicong and Ouyang, Liangyang and Lan, Fengbo and Wang, Zhixiang and Zhang, Kaipeng},
  journal={arXiv preprint arXiv:2610.31466},
  year={2026}
}

@article{ying2026wbench,
  title={Wbench: A comprehensive multi-turn benchmark for interactive video world model evaluation},
  author={Ying, Kaining and Hu, Hengrui and Ren, Siyu and Li, Jiamu and Chen, Fengjiao and Wang, Ziwen and Cao, Xuezhi and Cai, Xunliang and Ding, Henghui},
  journal={arXiv preprint arXiv:2605.25874},
  year={2026}
}

@article{wu2025qwenimage,
  title={Qwen-Image Technical Report},
  author={Wu, Chenfei and Li, Jiahao and Zhou, Jingren and Lin, Junyang and Gao, Kaiyuan and Yan, Kun and Yin, Sheng-ming and Bai, Shuai and Xu, Xiao and Chen, Yilei and others},
  journal={arXiv preprint arXiv:2508.02324},
  year={2025}
}

@article{oquab2024dinov2,
  title={{DINOv2}: Learning Robust Visual Features without Supervision},
  author={Oquab, Maxime and Darcet, Timoth{\'e}e and Moutakanni, Th{\'e}o and Vo, Huy and Szafraniec, Marc and Khalidov, Vasil and Fernandez, Pierre and Haziza, Daniel and Massa, Francisco and El-Nouby, Alaaeldin and others},
  journal={Transactions on Machine Learning Research},
  year={2024}
}

@inproceedings{xu2022vitpose,
  title={{ViTPose}: Simple Vision Transformer Baselines for Human Pose Estimation},
  author={Xu, Yufei and Zhang, Jing and Zhang, Qiming and Tao, Dacheng},
  booktitle={Advances in Neural Information Processing Systems},
  year={2022}
}

@article{hacohen2026ltx2,
  title   = {{LTX-2}: Efficient Joint Audio-Visual Foundation Model},
  author  = {HaCohen, Yoav and Brazowski, Benny and Chiprut, Nisan and Bitterman, Yaki and Kvochko, Andrew and Berkowitz, Avishai and Shalem, Daniel and Lifschitz, Daphna and Moshe, Dudu and Porat, Eitan and Richardson, Eitan and Shiran, Guy and Chachy, Itay and Chetboun, Jonathan and Finkelson, Michael and Kupchick, Michael and Zabari, Nir and Guetta, Nitzan and Kotler, Noa and Bibi, Ofir and Gordon, Ori and Panet, Poriya and Benita, Roi and Armon, Shahar and Kulikov, Victor and Inger, Yaron and Shiftan, Yonatan and Melumian, Zeev and Farbman, Zeev},
  journal = {arXiv preprint arXiv:2601.03233},
  year    = {2026}
}

@inproceedings{lin2026da3,
  title     = {Depth Anything 3: Recovering the Visual Space from Any Views},
  author    = {Lin, Haotong and Chen, Sili and Liew, Jun Hao and Chen, Donny Y. and Li, Zhenyu and Zhao, Yang and Peng, Sida and Guo, Hengkai and Zhou, Xiaowei and Shi, Guang and Feng, Jiashi and Kang, Bingyi},
  booktitle = {International Conference on Learning Representations},
  year      = {2026}
}

@article{keetha2025mapanything,
  title   = {{MapAnything}: Universal Feed-Forward Metric {3D} Reconstruction},
  author  = {Keetha, Nikhil and M{\"u}ller, Norman and Sch{\"o}nberger, Johannes and Porzi, Lorenzo and Zhang, Yuchen and Fischer, Tobias and Knapitsch, Arno and Zauss, Duncan and Weber, Ethan and Antunes, Nelson and Luiten, Jonathon and Lopez-Antequera, Manuel and Rota Bul{\`o}, Samuel and Richardt, Christian and Ramanan, Deva and Scherer, Sebastian and Kontschieder, Peter},
  journal = {arXiv preprint arXiv:2509.13414},
  year    = {2025}
}

@article{peng2026compression,
  title={Compression and Retrieval: Implicit Memory Retrieval for Video World Models},
  author={Peng, Zhan and Ma, Jie and Sun, Huiqiang and Gao, Chong and Xue, Zhijie and Pan, Zhiyu and Cao, Zhiguo and Liang, Jun and Li, Jing},
  journal={arXiv preprint arXiv:2606.23105},
  year={2026}
}

@article{deng2026actionrope,
  title={ActionRoPE: action as coordinates for spatially consistent control in planar-camera game world models},
  author={Deng, Zhiyang and Wu, Yuzhe and Lin, Zijun and Jin, Yeying},
  journal={arXiv preprint arXiv:2610.25861},
  year={2026}
}
}

\clearpage
\setcounter{page}{1}
\setcounter{section}{0}
\renewcommand{\theHsection}{supp.\arabic{section}}
\renewcommand{\theHsubsection}{\theHsection.\arabic{subsection}}
\maketitlesupplementary
\renewcommand{\thesection}{\Alph{section}}
\noindent
\begin{minipage}{\textwidth}
  \centering
  \includegraphics[
    width=\linewidth,
    height=0.40\textheight,
    keepaspectratio
  ]{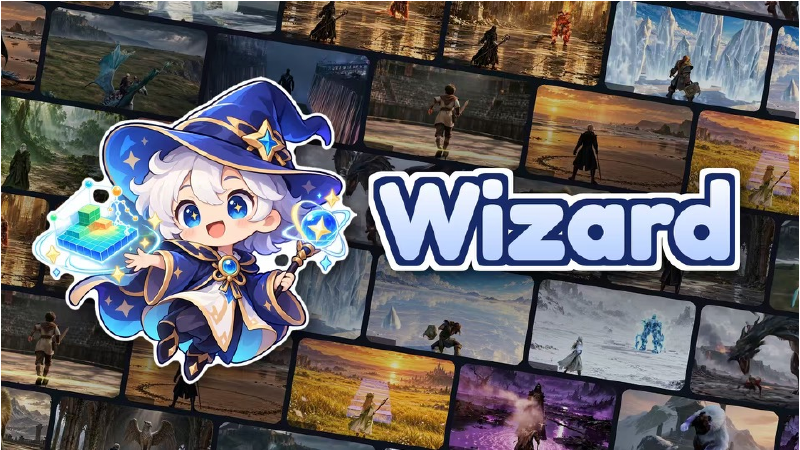}
\end{minipage}

\par\noindent
\begin{minipage}{\textwidth}

\begingroup
\setlength{\parindent}{0pt}
\setlength{\parskip}{0pt}

\vspace{10pt}
{\Large\bfseries Contents\par}
\vspace{2pt}

\newcommand{\wizardtocsection}[2]{%
  {\fontsize{10}{12}\selectfont\bfseries
    #1.\enspace #2\par}%
  \vspace{2pt}%
  \hrule height 0.3pt
  \vspace{3pt}%
}

\newcommand{\wizardtocentry}[2]{%
  \par\noindent
  \hangindent=2.6em
  \hangafter=1
  \makebox[2.6em][l]{\textbf{#1}}#2\par
}

\fontsize{9}{10.5}\selectfont

\noindent
\begin{minipage}[t][0.36\textheight][s]{0.48\linewidth}

  \vspace{0pt}

  \wizardtocsection{A}{Dataset}
  \wizardtocentry{A.1}{Data Sources}
  \wizardtocentry{A.2}{Unified Proxy Encoding}
  \wizardtocentry{A.3}{Filtering and Annotation}

  \vfill

  \wizardtocsection{B}{Implementation Details}
  \wizardtocentry{B.1}{Training Configuration}
  \wizardtocentry{B.2}{World-Coordinate Tokens}
  \wizardtocentry{B.3}{World Cache and Memory Retrieval}
  \wizardtocentry{B.4}{Few-Step Distillation}

  \vfill

  \wizardtocsection{C}{Benchmarks}
  \wizardtocentry{C.1}{Benchmark Design}
  \wizardtocentry{C.2}{WBench}
  \wizardtocentry{C.3}{Gameplay Set}
  \wizardtocentry{C.4}{Entry \& Revisit Set}
  \wizardtocentry{C.5}{Held-Out Revisit Windows}
\end{minipage}%
\hfill
\begin{minipage}[t][0.36\textheight][s]{0.48\linewidth}
  \vspace{0pt}

  \wizardtocsection{D}{Metrics}
  \wizardtocentry{D.1}{WBench Metrics}
  \wizardtocentry{D.2}{Gameplay Fidelity}
  \wizardtocentry{D.3}{Ablation Metrics}

  \vfill

  \wizardtocsection{E}{Baseline Settings}
  \wizardtocentry{E.1}{Model Versions}
  \wizardtocentry{E.2}{Control Conditioning}
  \wizardtocentry{E.3}{WBench Evaluation Protocol}

  \vfill

  \wizardtocsection{F}{Qualitative Results}
  \wizardtocentry{F.1}{Comparison with Baseline Methods}
  \wizardtocentry{F.2}{Qualitative Ablation}
\vfill
\wizardtocsection{G}{Limitations and Future Work}

\end{minipage}

\par
\endgroup
\end{minipage}
\par

\clearpage

\ifdefined\linenumbers
  \linenumbers
\fi
\raggedbottom

\section{Dataset}
\label{suppsec:data}

This section details the data summarized in Fig. \ref{fig:data} of the main paper, following its three stages: data acquisition (\cref{subsec:data_sources}), unified proxy encoding (\cref{subsec:data_process}), and filtering and annotation (\cref{subsec:data_prep}).

\subsection{Data Sources}
\label{subsec:data_sources}
Our training set contains 50,115 clips (56.1 hours) spanning 805 scenes and 178 characters. As shown in Fig. \ref{fig:data}, it comes from four complementary sources: gameplay recordings, public datasets, UE simulations and synthetic data. We detail each component in Tab. \ref{tab:supp_data}. These data sources offer complementary strengths: gameplay recordings provide high-quality appearance and motion but cover a limited range of characters and events; public datasets expand the diversity of creature morphologies and camera movements; and simulated and synthetic data fill gaps with events rarely captured in recordings, such as character entry and disappearance, as well as rapid camera turns.

\begin{table}[h]
  \centering
  \caption{\textbf{Overview of the training dataset.} We summarize the content, number of clips, duration, and proportion of data from each source. All training clips contain 121 frames at 30 fps, with proportions measured by duration.}
  \label{tab:supp_data}
  \setlength{\tabcolsep}{3pt}
  \renewcommand{\arraystretch}{1.15}
  \resizebox{\columnwidth}{!}{%
  \begin{tabular}{@{}llrrr@{}}
    \toprule
    \textbf{Source} & \textbf{Content} & \textbf{Clips} & \textbf{Hours} & \textbf{Share} \\
    \midrule
    \textbf{Gameplay recordings} & & \textbf{11,286} & \textbf{12.6} & \textbf{22\%} \\
    \quad Elden Ring          & combat, exploration    & 11,286 & 12.6 & \\
    \midrule
    \textbf{Public datasets} & & \textbf{26,055} & \textbf{29.2} & \textbf{52\%} \\
    \quad WildWorld           & creature combat        & 23,654 & 26.5 & \\
    \quad SceneFly            & camera fly-through     & 2,401  & 2.7 & \\
    \midrule
    \textbf{UE simulations} & & \textbf{2,966} & \textbf{3.3} & \textbf{6\%} \\
    \quad Unreal Engine       & scripted boss fights   & 2,966  & 3.3 & \\
    \midrule
    \textbf{Synthetic data} & & \textbf{9,808} & \textbf{11.0} & \textbf{20\%} \\
    \quad Procedural rendering & multi-character scenes & 9,386  & 10.5 & \\
    \quad LTX-2.5 generation   & world editing          & 422    & 0.5 & \\
    \midrule
    Total & & 50,115 & 56.1 & 100\% \\
    \bottomrule
  \end{tabular}%
  }
\end{table}

\paragraph{Gameplay Recordings (12.6 h).}
We record Elden Ring~\cite{fromsoftware2022eldenring}: 8.8 hours of combat against six bosses with diverse body structures and 3.8 hours of exploration across 11 regions, with more than 20 player outfits.
Camera poses and skeletons are read from the game process; scene depth is captured once per second, fused into scene geometry and re-rendered at every frame.

\paragraph{Public Datasets (29.2 h).}
WildWorld~\cite{li2026wildworld} adds diverse creature morphologies; its annotations are converted to our proxy representation.
SceneFly~\cite{peng2026compression} provides character-free camera fly-throughs that return to earlier viewpoints, which supervises scene memory; its depth is estimated with MapAnything~\cite{keetha2025mapanything}.

\paragraph{UE Simulations (3.3 h).}
We script boss fights in Unreal Engine with reconstructed characters and weapons~\cite{tochilkin2024triposrfast3dobject}.
Since the engine state is known, the skeletons, cameras and scene geometry of these clips are exact.

\paragraph{Synthetic Data (11.0 h).}
This source has two parts.
First, we procedurally render 162 character assets (133 bosses and 29 players) from Sketchfab~\cite{sketchfab}, Quaternius~\cite{quaternius} and KayKit~\cite{kaykit} in scenes built from 742 Poly Haven~\cite{polyhaven} scene assets.
The clips cover four scenarios that recordings rarely contain: novel-character entry, multi-character coexistence, disappearance and reappearance, and rapid turns under a following camera.
All conditions come directly from the renderer.
Second, for world editing we generate videos with LTX-2.5~\cite{hacohen2026ltx2} in which the terrain in front of the player is reshaped (\eg, stairs, pillars, walls) or effects with clear geometry appear (\eg, water columns, ice spikes).
Their conditions are estimated with Depth Anything 3~\cite{lin2026da3}, SAM~2~\cite{ravi2025sam} and ViTPose~\cite{xu2022vitpose}.

\subsection{Unified Proxy Encoding}
\label{subsec:data_process}
Every source is converted to the same visual proxy through the two branches of Fig. \ref{fig:data}. In the \textit{scene branch}, characters are removed from the depth and the region behind them is completed, which gives the scene geometry.
In the \textit{character branch}, the 3D joint positions of each character are drawn as a skeleton. Both are rendered into one conditioning video under the shared camera.

Following Marionette~\cite{meng2026marionette}, the RGB channels encode geometry rather than color: $\mathrm{R}$ encodes the height and $\mathrm{B}$ the camera-forward depth of both the scene and the characters. The $\mathrm{G}$ channel has a different meaning in the two branches. For characters, it gives the slot of each character, separating the player from the others, and its facing direction. For the scene, it gives how reliable the geometry is: unknown regions take a reserved dark green, coarse geometry a constant level, and exact geometry a brighter level with Lambertian shading. The proxy does not indicate which reference image a character corresponds to; this binding is left to the character-aware adaptive mask.

\subsection{Filtering and Annotation}
\label{subsec:data_prep}
Sequences are split into non-overlapping 121-frame windows at 30 FPS, resized to $832 \times 480$ (references to $480 \times 480$), and encoded with the Wan2.2 VAE.
Invalid clips, \ie, those whose state information is incomplete or whose proxy does not align with the video, are discarded.
Qwen3-VL-8B~\cite{bai2025qwen3} then writes a caption for each remaining clip from its first frame and character references, which yields the aligned proxy--video pairs used for training.

\section{Implementation Details}
\label{sec:supp_impl}

\subsection{Training Configuration}
We initialize from Wan2.2-TI2V-5B~\cite{wan2025wan} and keep its VAE and UMT5 text encoder~\cite{chung2023unimax} frozen.
The bidirectional model is trained for 35K steps on 8 NVIDIA H200 GPUs with one clip per GPU.
We use AdamW with a learning rate of $10^{-5}$ for the pretrained DiT weights and $10^{-4}$ for the new conditioning modules, and drop text with probability 0.1.
At inference, the bidirectional model uses 30 denoising steps and a guidance scale of 3.

\subsection{World-Coordinate Tokens}
Each location of the proxy and memory frames carries its world coordinate, a flag for whether it has geometry, and the geometry level of the proxy.
Coordinates are encoded with sinusoids at 8 frequencies and multiplied by the flag, so unknown regions, sky and characters contribute no coordinates.
The features are packed to the video-token resolution, mapped by a zero-initialized linear layer, and added to the patch embeddings of the video and memory tokens.
Characters are hidden in memory frames, so that memory supplies the scene only.

\subsection{World Cache and Memory Retrieval}
At inference, the proxy is rendered from the world cache at every frame.
Generated keyframes are lifted into 3D with MapAnything~\cite{keetha2025mapanything} and fused into the cache by a background worker, so generation never waits for reconstruction.
For each chunk of 4 latent frames, up to three memory frames are selected greedily from earlier frames by how much of the currently visible region each newly covers.
The same retrieval function builds the training data, so the model is trained on the kind of memory it receives at inference.

\subsection{Few-Step Distillation}
\label{subsec:supp_distill}

\paragraph{Training Stages.}
Following Causal Forcing++~\cite{zhao2026causal}, we obtain the causal model in three stages.
All stages read a sliding window of 20 latent frames plus the first frame as a persistent attention sink.

\noindent\textit{Causal adaptation} (11K steps). The bidirectional model is trained with teacher forcing and independent per-frame noise levels.

\noindent\textit{Consistency distillation} (10K steps). A guided causal teacher takes one Euler step along a 48-level discretization, and the student matches the prediction of its EMA at the next level.

\noindent\textit{Distribution matching distillation} (480 steps). The student generates clips autoregressively from its own history and is updated with the score difference between the frozen bidirectional model and a critic trained on the student's samples.

\paragraph{History Corruption.}
Under teacher forcing, the clean history already contains almost all of the scene, so the model does not learn to rely on its memory when the history drifts at inference.
In the first two stages we therefore replace the clean history, with probability $p$, by the model's one-step reconstruction of a noised copy with per-frame noise level $s \sim \mathcal{U}(0, s_{\max})$, while the sink frame, the memory frames and the target stay clean.
We use $p{=}0.5$, $s_{\max}{=}0.25$ in causal adaptation and $p{=}0.7$, $s_{\max}{=}0.5$ in consistency distillation.
The corruption is applied only to clips that require memory (revisits and SceneFly).

\paragraph{Deployment.}
We use the EMA student after distillation. It streams with 2 denoising steps and a KV cache of 20 latent frames plus the sink frame.

\section{Benchmarks}
\label{sec:supp_bench}

\subsection{Benchmark Design}

Existing world-model benchmarks such as WBench~\cite{ying2026wbench} evaluate visual quality, navigation and consistency when revisiting a view, with a single controllable subject.
They do not test whether new characters can be brought into the world, whether each one keeps its specified identity, or whether scripted interactions and their consequences actually take place.
We therefore keep WBench for visual quality and navigation and add three test sets for these abilities (\cref{tab:supp_bench}).
All three follow the same principles, which also underlie the fairness of the comparison in \cref{sec:supp_metrics}.

\noindent\textbf{Unseen Content.} All sets are built on WBench first frames, and all characters are specified by newly generated reference images; neither is seen during training.

\noindent\textbf{Engine Ground Truth.} Every case is produced by our engine from a script, so the position and state of each character at each frame are known exactly. Cases are filtered only by engine-side checks, independently of any generated video.

\noindent\textbf{Model-Agnostic.} The same script is rendered into every control format (text, skeleton video, depth and semantic maps), so that all methods receive equivalent control instructions through inputs they are designed to support.

\begin{table}[t]
  \centering
  \caption{\textbf{Test Sets.} WBench evaluates navigation; our three sets evaluate multi-character gameplay, identity binding and memory.}
  \label{tab:supp_bench}
  \setlength{\tabcolsep}{3pt}
  \renewcommand{\arraystretch}{1.15}
  \resizebox{\columnwidth}{!}{%
    \begin{tabular}{@{}lcll@{}}
      \toprule
      \textbf{Set} & \textbf{Cases} & \textbf{Tests} & \textbf{Metrics} \\
      \midrule
      WBench navigation & 158 & visual quality, navigation & official \\
      Gameplay          & 50  & scripted multi-character combat & interaction adh., mechanics cons. \\
      Entry \& Revisit  & 69  & identity binding, scene memory & reference binding, spatial cons. \\
      Held-out revisit  & 100 & memory under large view change & revisit error, copy bias \\
      \bottomrule
    \end{tabular}%
  }
\end{table}

\subsection{WBench}
We evaluate the performance of baseline methods and Wizard on all 158 cases, comprising 74 third-person and 84 first-person scenarios, following the official evaluation protocol for both perspectives.

\subsection{Gameplay Set}
The set contains 50 clips of 121 frames on 25 third-person first frames from WBench.
Characters are specified by reference images generated with Qwen-Image~\cite{wu2025qwenimage} (humanoids, giants and quadrupeds), automatically rigged, and animated with combat motions from Elden Ring~\cite{fromsoftware2022eldenring} and WildWorld~\cite{li2026wildworld}.
In each clip, two characters enter from both sides (35 humanoid and 15 quadruped pairs) and follow one of four scripts, which cover the basic mechanics of a combat game: the player attacks a designated character (13 clips); a character attacks and the player evades (13); a character is knocked down and stays down (12); two characters attack the player in turn (12).
Every script involves three characters and requires a specific one to act, so a model succeeds only if it introduces both characters, keeps them distinct, and executes the interaction between the right pair.

\subsection{Entry \& Revisit Set}
The set is built in the same way on 69 first frames (56 third-person and 13 first-person).
Each first frame has two sequences, and each sequence isolates one of our two modules.
In the entry sequence (121 frames), the two entering characters share one skeleton and differ only in their reference images, so the control signal carries no identity cue and the characters can only be told apart through reference binding.
In the revisit sequence (241 frames), the camera turns away by $150^\circ$ and returns, so the original view leaves the frame entirely and must be restored from memory.

\subsection{Held-Out Revisit Windows}
The world-coordinate ablation needs ground truth for the revisited content, which generated sequences do not have.
It therefore uses 100 windows of 121 frames from SceneFly dataset \cite{peng2026compression} of three scenes held out from training, with ground-truth depth, camera poses and video.
To test memory under a large view change, the memory frames of each window are chosen by geometry from other windows of the same scene: a frame qualifies if it covers at least 10\% of the current view and its camera is rotated by 45--60$^\circ$ from it.
Both variants receive the same memory frames, seed and sampling steps.

\section{Metrics}
\label{sec:supp_metrics}

We use three groups of metrics: the official WBench metrics for visual quality and spatiotemporal quality, two gameplay-fidelity metrics for the comparison with baselines, and four metrics for the ablations.

\subsection{WBench Metrics}
All metrics follow the official WBench~\cite{ying2026wbench} implementation.
For spatial consistency we report the gated version, which compares the frames revisiting the initial view with the first frame and penalizes videos whose camera does not move as instructed.

\subsection{Gameplay Fidelity}
\label{subsec:supp_gameplay_metrics}
The WBench metrics score how a video looks, not what happens in it: a video in which no character enters, or the wrong character attacks, can still obtain high quality and consistency scores.
We therefore measure gameplay fidelity with two metrics.
Both are defined as the accuracy of closed-form questions whose correct answers are given by the engine state, so they require neither a subjective rating nor a reference video.

\paragraph{Protocol.}
Every query is answered by GPT-5.5~\cite{openai2026gpt55} and consists of (i) frames sampled from the generated video at times fixed by the script, (ii) a crop of the player from the first frame, (iii) a text description of each of the two characters, and (iv) one multiple-choice question whose options are shuffled.
Let $\mathcal{Q}$ be a set of queries, $a_q$ the judge's answer to query $q$, and $a_q^{*}$ the answer given by the engine state.
Each metric is an accuracy
\begin{equation}
  \mathrm{Acc}(\mathcal{Q}) = \frac{1}{|\mathcal{Q}|} \sum_{q \in \mathcal{Q}} \mathbb{1}\!\left[a_q = a_q^{*}\right],
\end{equation}
pooled over all queries of the 50 clips.

\noindent\textbf{Interaction Adherence} measures whether the specified character performs the intended action in the generated rollout. Its query set contains one query for each of the 75 scripted actions. For an action, we sample 8 frames around its scripted time and ask who is involved, \eg, ``\textit{Which character attacks the player character?}'' or ``\textit{Which character rolls or dodges away?}'', with the candidate characters and ``nobody'' as options. The correct answer is the character that the script assigns to the action. An answer is incorrect if the intended action is absent or performed by the wrong character. Generating an arbitrary attack is therefore insufficient: the model must show both characters, preserve their distinct identities, and have the designated character perform the specified action.

\noindent\textbf{Mechanics Consistency} measures whether the state of the world follows the script over time, inspired by StatePlay~\cite{lin2026stateplay}.
It is the mean of two accuracies,
\begin{equation}
  \mathrm{MC} = \tfrac{1}{2}\left(\mathrm{Acc}(\mathcal{Q}_{\text{state}}) + \mathrm{Acc}(\mathcal{Q}_{\text{event}})\right).
\end{equation}

\textit{State Persistence} ($\mathcal{Q}_{\text{state}}$, 352 queries) measures whether the generated rollout preserves the engine-maintained character states, including their presence, number, and defeat status. In temporal windows away from entry and knock-down events, the judge determines whether each character is absent, standing, or lying defeated, and whether it appears more than once. An answer is correct if it matches the character's state in the engine at that time. This penalizes missing characters, unintended disappearances, duplicates, and defeated characters standing up again.

\textit{Event Realization} ($\mathcal{Q}_{\text{event}}$, 112 queries) measures whether the generated rollout follows game rules by realizing their expected visual consequences, such as a character falling defeated when its HP reaches zero. After each entry event, the judge checks whether the entering character appears in the scene; after each knock-down event, the judge checks whether the defeated character is lying on the ground. This verifies that rule-driven state transitions are reflected in the generated video.

\subsection{Ablation Metrics}
\label{subsec:supp_abl_metrics}
The ablations compare variants of our own model that receive identical inputs, seeds and sampling steps, so each metric below isolates the effect of one component.
The module ablation uses two metrics on the Entry \& Revisit set, and the world-coordinate ablation two metrics on the held-out windows.

\noindent\textbf{Reference Binding} measures whether each entering character takes the appearance of its own reference image.
For a frame $t$ in which character $c \in \{1,2\}$ is visible, let $x_{t,c}$ be the crop of the generated frame at the box of $c$, and $r_1, r_2$ the two reference images.
With $\phi$ the DINOv2~\cite{oquab2024dinov2} feature and $\bar{c}$ the other character,
\begin{equation}
\begin{aligned}
  \mathrm{RB}
  = \frac{1}{|\mathcal{T}|}
  \sum_{(t,c) \in \mathcal{T}}
  \mathbb{1}\!\Big[
  &\cos\!\big(\phi(x_{t,c}), \phi(r_c)\big) \\
  &> \cos\!\big(\phi(x_{t,c}), \phi(r_{\bar{c}})\big)
  \Big],
\end{aligned}
\end{equation}
where $\mathcal{T}$ contains the frames, sampled at 2 fps, in which the character covers at least 0.2\% of the image.
The boxes come from the engine and are the same for every variant, so the metric does not depend on a detector.
Because the two characters of an entry sequence share one skeleton, a model that ignores the references, or gives both characters one appearance, scores the chance level of 50\%.

\noindent\textbf{Spatial Consistency} in the ablation study is evaluated using the metrics defined in WBench. The scores in Tab.~\ref{tab:consistency_ablation} differ from those in Tab.~\ref{tab:main_results} because the ablation uses a subset of revisit-intensive cases to assess how spatial memory preserves scene consistency when previously observed regions are revisited.

\noindent\textbf{Revisit Error} measures how accurately remembered content is restored from different viewpoints.
On the held-out windows the ground-truth video $x$ is available.
Let $\Omega$ be the \textit{memory region}, \ie, the pixels that are covered by a memory frame but not visible in the first frame of the window; these pixels can only be restored from memory.
With $\hat{x}$ the output,
\begin{equation}
  \mathrm{RE} = \frac{1}{|\Omega|} \sum_{p \in \Omega} \big|\hat{x}(p) - x(p)\big|,
\end{equation}
in intensity levels (0--255); lower is better.
$\Omega$ is computed from the ground-truth depth and cameras only, so it is identical for both variants and independent of their outputs. We introduce a new metric because the trajectories revisit previously observed regions from different viewing angles, rather than returning to the same viewpoints required by the spatial consistency metric. This metric therefore evaluates scene consistency across revisits with viewpoint changes.

\noindent\textbf{Copy Bias} evaluates whether remembered regions are correctly re-rendered from a new viewpoint rather than directly copied from the memory frame, helping assess the benefit of world-coordinate conditioning.
Let $M$ denote the raw memory frame without reprojection, $x$ the ground-truth current view, and $\hat{x}$ the generated view. We define
\begin{equation}
  \mathrm{CB} = \left| \mathrm{L1}_\Omega(x, M)
  - \mathrm{L1}_\Omega(\hat{x}, M) \right|,
\end{equation}
where $\mathrm{L1}_\Omega(a,b)$ is the mean absolute difference over the revisited region $\Omega$.
The first term measures the expected visual difference between the current view and the memory frame, while the second measures the corresponding difference for the generated view.
Directly copying $M$ makes the second term zero, yielding a bias equal to the first term. Conversely, excessive deviation from $M$ can make the second term larger than the first, which is also penalized.
A perfect output yields zero bias; lower is better.
We evaluate viewpoint changes of 45--60$^\circ$, where direct copying and correct re-rendering produce clearly different appearances.

\section{Baseline Settings}
\label{sec:supp_baselines}

\subsection{Model Versions}
We use the released checkpoints: YUME 1.5 (5B, 720p), Matrix-Game 3.0 (base model), LingBot-World-Base (camera-controlled), Marionette's observation model on Wan2.2-Fun-5B-Control, and CWM's adapter on MiniMax-H3.
Model sizes and training data in \cref{tab:main_results} are taken from the original papers and exclude base-model pretraining.
The \textit{Steps} column lists the sampling steps of our runs: 50 for YUME and Matrix-Game, and the official defaults for Marionette (40), LingBot-World (70) and CWM (20).

\subsection{Control Conditioning}
All models start from the same first frames and receive the control signals through their supported inputs.
Text-conditioned baselines share one prompt per clip: Qwen3.6-27B describes the scene, the player and each character, and the scripted events are written with their times.
CWM~\cite{chen2026code} is run in its official setting, with depth and semantic maps rendered from the same scripts and a prompt in its released format.
Marionette~\cite{meng2026marionette} is conditioned on rendered skeleton videos.
Only Wizard receives reference images; as described in \cref{subsec:supp_gameplay_metrics}, the judge does not see them.

\subsection{WBench Evaluation Protocol}
Results of YUME, Matrix-Game and LingBot-World are taken from the official leaderboard, which evaluates the same releases; the other methods are evaluated by us with the official code.
For the 16-second videos, bidirectional models (CWM, Marionette and Wizard) generate window by window, each conditioned on previously generated frames, and our causal model generates a continuous stream.

\section{Qualitative Results}
\label{sec:supp_qual}

\subsection{Comparison with Baseline Methods}
\begin{figure*}[t]
  \centering
  \includegraphics[width=\textwidth]{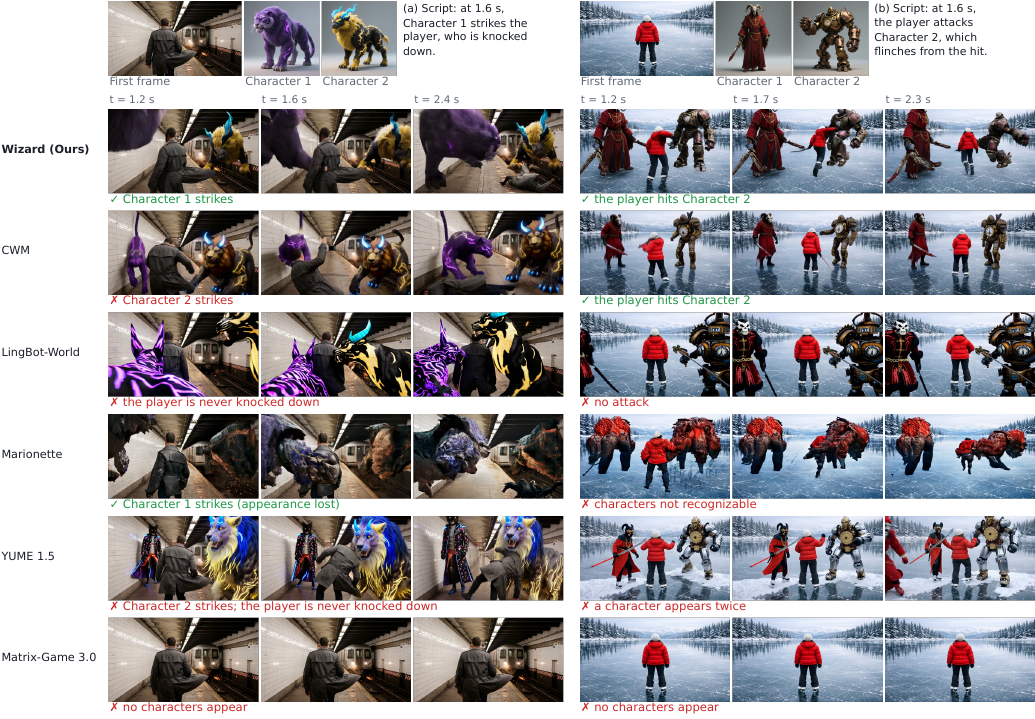}
  \caption{\textbf{Qualitative Comparison with the Baseline Methods.} Each case shows the first frame, the two character references and the script; frames are taken at the same times for all methods, and the line below each row gives the judge's outcome.}
  \label{fig:qual_cmp}
\end{figure*}

\cref{fig:qual_cmp} compares all methods on two Gameplay cases.
In (a), the script asks Character 1 to strike the player: Wizard follows it, in CWM and YUME the other character attacks, and in LingBot-World and YUME the player is never knocked down.
Marionette performs the right action but loses the appearance of both characters.
In (b), the player attacks Character 2: Wizard and CWM both execute the attack, whereas no attack happens in LingBot-World, the characters are not recognizable in Marionette, and one character appears twice in YUME.
Matrix-Game accepts only keyboard and mouse actions, so no character enters.

\subsection{Qualitative Ablation}
\begin{figure*}[t]
  \centering
  \includegraphics[width=\textwidth]{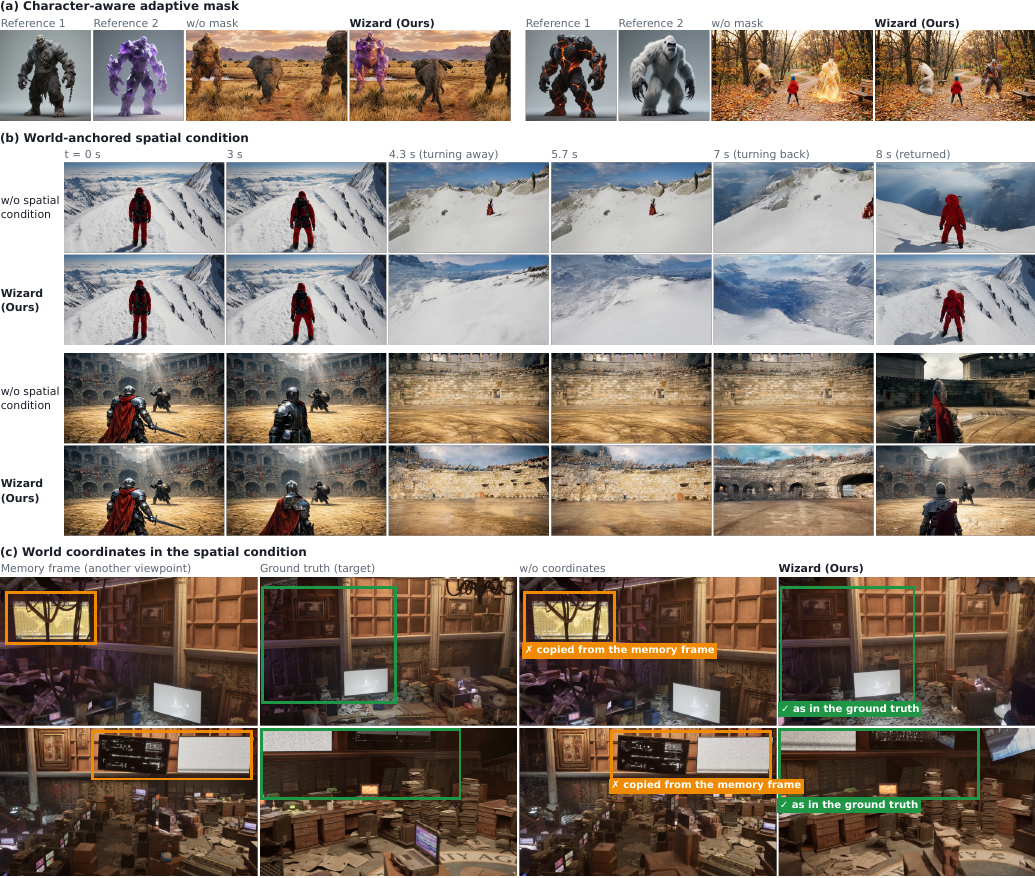}
  \caption{\textbf{Qualitative Ablation.} (a) Entry sequences without and with the mask. (b) Revisit sequences without and with the spatial condition: the camera turns away and returns. (c) Held-out windows without and with world coordinates; orange boxes mark content that matches the memory frame, and green boxes mark the same area rendered as in the ground truth.}
  \label{fig:abl_qual}
\end{figure*}

\begin{figure*}[t]
  \centering
  \includegraphics[width=\textwidth]{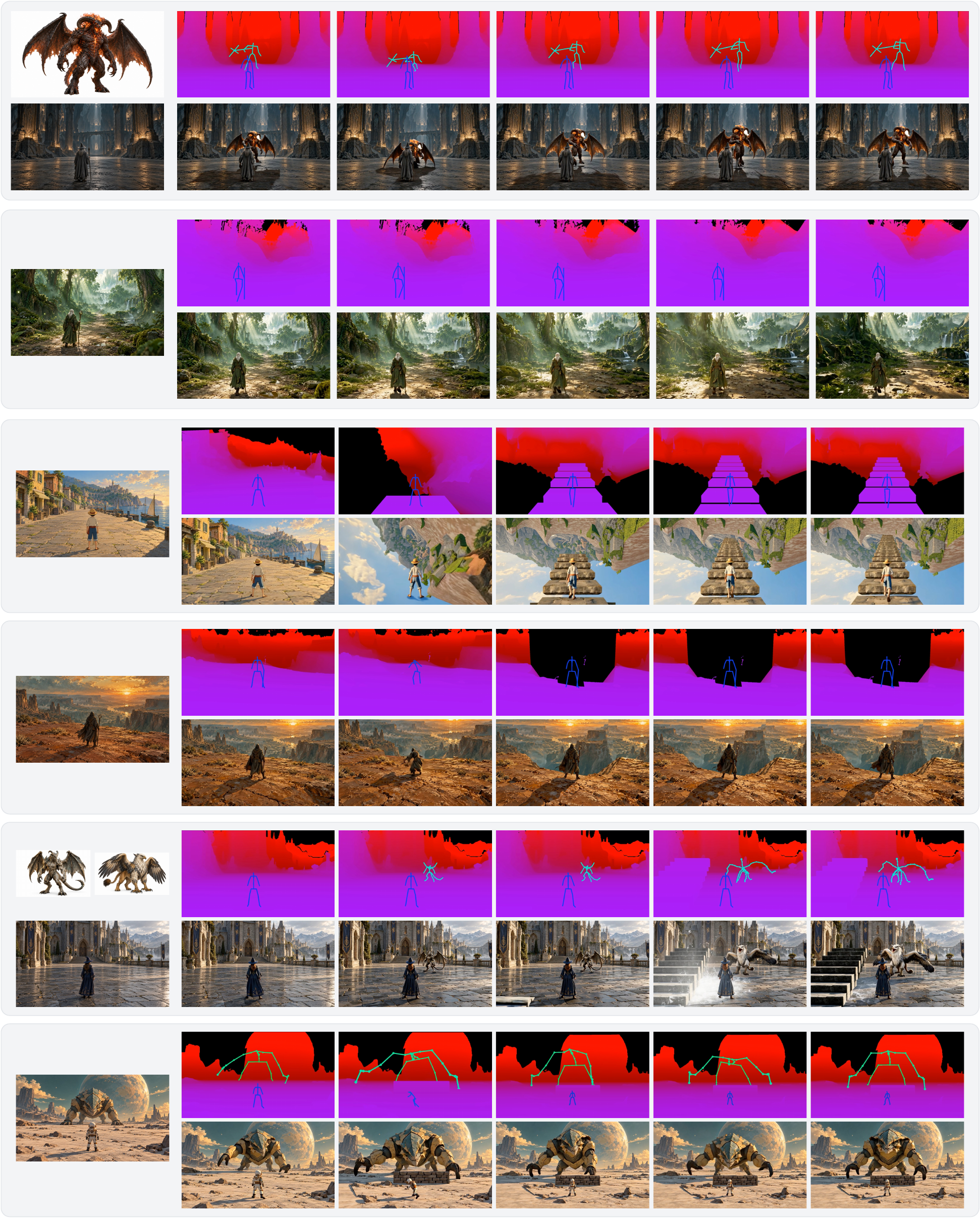}
  \caption{\textbf{More Qualitative Examples.} For each example, reference images are shown on the left, with visual proxies in the top row and the corresponding video rollouts in the bottom row. Please zoom in for better visibility.}
  \label{fig:more_visual}
\end{figure*}

\cref{fig:abl_qual} illustrates the effect of each component.
(a) Without the character-aware adaptive mask, the two entering characters take a mixture of both references or share one appearance; with the mask, each character follows its own reference.
(b) Without the world-anchored spatial condition, the model generates a different scene when the camera returns; with it, the original view is restored.
(c) The memory frame shows the scene from a viewpoint 45--60$^\circ$ away from the current one. The variant without world coordinates reproduces the content of the memory frame at its original image position (orange box), whereas the full model re-renders it under the current view, close to the ground truth.
Additional qualitative examples are provided in \cref{fig:more_visual}.

\section{Limitations and Future Work}
\label{sec:supp_limit}
Wizard still has several limitations.

\noindent\textbf{Motion coverage.} Character motion is retrieved from a motion library and retargeted, so it is difficult to find compatible motions for characters with uncommon body structures.
Generating motion directly for arbitrary skeletons would remove this dependence.

\noindent\textbf{Long rollouts.} The streaming model reads a bounded history and relies on retrieved memory frames for everything older, so the consistency between the scene proxies and the generated video may decline over longer rollouts.

\noindent\textbf{Upstream perception.} The world state is built with off-the-shelf models for rigging and 3D reconstruction, and their errors propagate: a wrongly rigged limb or a misplaced surface produces a proxy that the generator follows faithfully.
Correcting the world state from the generated frames could reduce this sensitivity.

Future work will focus on more flexible and realistic character motion that better reflects actual gameplay, consistent scene appearance over long rollouts, and richer interactions among more characters.
These improvements will bring agentic video generation closer to fully playable, closed-loop game worlds with greater player freedom and more precise, nuanced control.

\end{document}